\documentclass{article}

\usepackage[utf8]{inputenc}
\usepackage{my_stylefile}
\usepackage{anyfontsize}

\usepackage{comment}
\usepackage{natbib}
\def\XX{\textrm{\textsc{Alrite}}}
\usepackage{fancyhdr}       

\begin{document}

\fancyhead[LO]{Asymmetrical Latent Representation For ITE Modeling}

\title{Asymmetrical Latent Representation for Individual Treatment Effect Modeling
}

\author{
  Armand Lacombe, Michèle Sebag \\
  AO/TAU, CNRS - INRIA \& LISN \& Univ. Paris-Saclay \\
  \texttt{\{name\}@lisn.fr} \\
}

\maketitle

\begin{abstract}
{Conditional Average Treatment Effect (\CATE) estimation, at the heart of counterfactual reasoning, is a crucial challenge for causal modeling both theoretically and applicatively, in domains such as healthcare, sociology, or advertising. 
Borrowing domain adaptation principles, a popular design maps the sample representation to a latent space that balances control and treated populations while enabling the prediction of the potential outcomes. This paper presents a new \CATE\ estimation approach based on the asymmetrical search for two latent spaces called {\it{\xx}} (\XX), where the two latent spaces are respectively intended to optimize the counterfactual prediction accuracy on the control and the treated samples. 
Under moderate assumptions, \XX\ admits an upper bound on the precision of the estimation of heterogeneous effects (\PEHE), and the approach is empirically successfully validated compared to the state of the art.}
\end{abstract}



\section{Introduction}

In the Neyman-Rubin framework \cite{rubinCausalInferenceUsing2005}, causal inference focuses on a simple question: how different would the outcome have been if the treatment had been different? Answering this question, however, raises considerable difficulties, as the true answer is inevitably unknown: the $i$-th individual either is treated (with outcome $y_i^1$) or not (with outcome $y_i^0$). In both cases, observation of the difference $y_i^1 - y_i^0$ remains inaccessible. Overall, the estimation of the individual causal effect, acknowledged as the \textit{fundamental problem of causal inference} \citep{hollandStatisticsCausalInference1986}, addresses a machine learning problem with missing values. 

A most natural approach consists in independently modeling outcome $Y^1$ (respectively $Y^0$) from the treated (resp. untreated, or control) samples as functions of their covariate description $X$, allowing the individual causal effect to be estimated from the difference of both models. This approach is valid subject to a key assumption, stating that treated and control samples are drawn from the same distribution $\PP(X)$; in other words, the treatment assignment must follow the randomized control trial methodology. 

In real-life applications however, this assumption rarely holds true. In the medical field the treatment variable $T$ is generally assigned by the doctor, depending on the severity of the individual's condition. In the field of education, the decision of e.g. following a curriculum depends on the individual's background. In such cases, the treated and control distributions $\PP_1(X) = P(X | T=1)$ and $\PP_0(X) = \PP(X|T=0)$ are different.

Estimating the potential outcome models based on the joint distributions $\PP(X,Y^1)$ and $\PP(X,Y^0)$ can thus be cast as a learning across domain problem \citep{ben-davidAnalysisRepresentationsDomain2006,ganinDomainAdversarialTrainingNeural2016}, where the two support distributions $P(X | T=0)$ and $P(X | T=1)$ differ. 
Building upon the state of the art \citep{ben-davidAnalysisRepresentationsDomain2006}, it thus comes naturally to integrate the space of original covariates into a latent space aligning both support distributions \citep{johanssonLearningRepresentationsCounterfactual2016a,shalitEstimatingIndividualTreatment2017} (more in \cref{chapter:sota}). 

\begin{figure*}[ht]
\centering
\includegraphics[width=.9\textwidth]{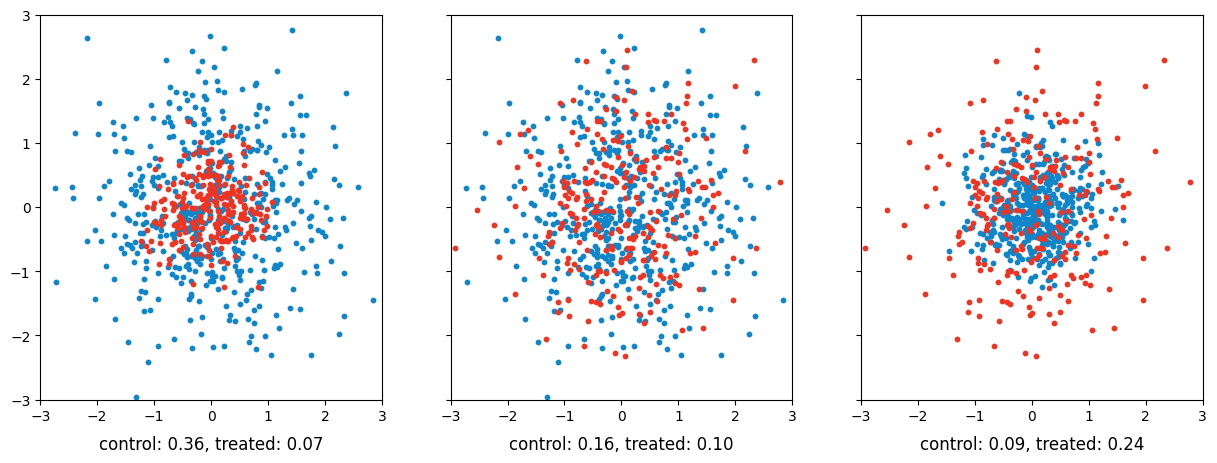}
\caption{Average counter-factualizability (definition in section \ref{subsub:counterfactualizability}) for \textcolor{C2}{treated} and \textcolor{B4}{control} samples in latent space. Left: the distributions support a good counter-factual estimation for treated samples (but less so for control samples). Middle: the distributions support a decent ounter-factual estimation for both populations. Right: the distributions support a good counter-factual estimation for control samples (but less so for treated samples).}
\label{fig:2DA}
\end{figure*}

A main claim of the paper is that the \CATE\ problem however significantly differs from the mainstream learning across domain setting. Specifically, estimating $\PP(Y^1|X).\PP_0(X)$ from $\PP(Y^1|X).\PP_1(X)$ requires all control samples to be `sufficiently close' from some treated samples (A). Likewise, estimating $\PP(Y^0|X).\PP_1(X)$ from $\PP(Y^0|X).\PP_0(X)$ requires all treated samples to be `sufficiently close' from some control samples (B). The difference between these requirements is illustrated on Fig. \ref{fig:2DA}: the leftmost and rightmost images correspond to criteria (A) and (B) respectively, while the center image corresponds to a representation suitable for learning across domains.

The main contribution of the present paper is to formalize the intuition that \CATE\ estimation involves two intertwined 
learning across domain
problems, appealing to different requirements.
The presented analysis motivates the design of an original model architecture, called {\it{\xx}} (\XX), as well as new learning criteria, offering formal guarantees on the quality of the learned \CATE\ estimate. 

This article is organized as follows. After presenting the formal background, \cref{chapter:sota} briefly introduces and discusses related work.  \cref{chapter:alrite} gives an overview of the proposed \XX, and presents its formal analysis. 
\cref{sec:resu} presents a comprehensive comparative empirical evaluation of the approach, and the paper concludes with some research perspectives. 

\paragraph*{Notations.} The observation dataset is noted $\mathcal{D} = \{ (x_i,t_i,y_i), i=1 \ldots n\}$, where $x_i \in \mathcal{X} \subset \RR^d$, $t_i \in \{0,1\}$ and $y_i \in \mathcal{Y} \subset \RR$ respectively denote the covariate, the treatment assignment and the observed outcome for the $i$-th sample, with $t_i=1$ (respectively $t_i=0$) for a treated (resp. control) sample; $n_0$ and $n_1$ respectively denote the number of control and treated samples. As usual, upper case letters denote random variables in the following, while lower case ones refer to observations.

\paragraph*{Assumptions. \label{def:positivity}} \XX\ is based on the three standard assumptions for counterfactual estimation \citep{rubinBayesianInferenceCausal1978, rosenbaumCentralRolePropensity1983} within the Neyman-Rubin framework \citep{rubinCausalInferenceUsing2005}: conditional exchangeability,\footnote{also referred to as \textit{exchangeability}  \citep{wuBetaIntactVAEIdentifyingEstimating2021}, \textit{weak unconfoundedness} \citep{hiranoPropensityScoreContinuous2005}, or \textit{conditional independence} in the econometrics field \citep{lechnerEarningsEmploymentEffects1999, angristMostlyHarmlessEconometrics2009}.} positivity,\footnote{also referred to as \textit{Overlap}.}, and  Stable Unit Treatment Value Assumption (SUTVA)\footnote{SUTVA is decomposed into "no spill-overs" and "consistency".}:

\begin{itemize}[leftmargin=120pt]
    \item[\sc{Conditional Exchangeability}:] $\forall t \in \{0,1\},\ Y^t \indep T \ |\ X$

    \item[\sc{Positivity}:] $\forall\ A \subset \mathcal{X}\ ,\ \PP(X \in A)>0\ \implies \PP(X \in A|T=1), \PP(X \in A|T=0) >0$

    \item[\sc{SUTVA}:] $Y = (1-T)\,Y^0+T \, Y^1$ 
\end{itemize}

\paragraph*{Quantity of interest and performance metrics.}
The Conditional Average Treatment Effect (\CATE, denoted by $\tau$) is an estimate of the expected impact of the treatment at the subpopulation/individual\footnote{See \cite{vegetabileDistinctionConditionalAverage2021a} for further developments about the distinction between Individual Treatment Effect and Conditional Average Treatment Effect.} level:
\begin{equation*}
    \tau:x \mapsto \EE[Y^1 - Y^0|X=x]
\end{equation*}
The performance of the \CATE\ estimate is usually assessed using the {Precision in Estimation of Heterogeneous Effect} (\PEHE) when the ground truth $\tau$ is available, and either the {Policy Risk} (\Rpol) or the {Observational Policy Risk} (\oRpol) otherwise, with:
\begin{alignat}{3}
    \PEHE : \htau \mapsto&\ \EE\big[(\tau(X)-\htau(X))^2\big] \nonumber\\
    \Rpol : \htau \mapsto&\ 1- \PP(\htau(X)>0)\EE[Y^1|\htau(X)>0]
    &&- \PP(\htau(X)\leq 0)\EE[Y^0|\htau(X)\leq 0] \nonumber\\
    \oRpol: \htau \mapsto&\ 
    1- \PP(\htau(X)>0)\EE[Y|\htau(X)>0,T=1] &&-\PP(\htau(X)\leq0)\EE[Y|\htau(X)\leq0,T=0]  
\label{eq:metricsdef}
\end{alignat}

Note that if the treatment assignment is uniform, then \Rpol\ and \oRpol\ coincide. For the sake of readability and when no confusion is to fear the empirical counterparts of the above expectations are still denoted \PEHE, \Rpol\ and \oRpol.

\section{State of the art \label{chapter:sota}}
As large datasets become increasingle available,  they support the estimation of causal effects at individual or group level, reflecting the heterogeneity of treatment impacts. The state of the art commonly distinguishes several categories of \CATE\ learners, depending on the main features of the models and notably how these models account for the treatment assignment variable \cite{kunzelMetalearnersEstimatingHeterogeneous2019, nieQuasioracleEstimationHeterogeneous2021,kennedyOptimalDoublyRobust2023}. 

\subsection{\slearners\ \label{par:slearners}} 
S(ingle)-learners 
{fit a single model with treatment assignment $T$ and covariates $X$ as inputs}. The potential outcome models are learned along classical supervised learning, with:
$$\hmu(x,t) \approx \mu^t(x) = \EE[Y|X=x, T=t]$$ 
and the treatment effect is defined as: $\htau(x) = \hmu(x,1) - \hmu(x,0)$.
In the \textit{Balancing Neural Network} (\textbf{BNN}) approach \cite{johanssonLearningRepresentationsCounterfactual2016a}, an embedding $\phi$ is used to map the covariate space onto a latent representation, merging the (image of) control and treatment distributions along the domain adaptation principles \citep{ben-davidAnalysisRepresentationsDomain2006}. The potential outcome $\hmu(x,t)$ is sought as a single neural net, operating on the concatenation of  $\phi(x)$ and the treatment variable $t$. 
However, the fact that the treatment assignment variable is given no particular role tends to bias the \CATE\ estimate toward 0 according to \cite{kunzelMetalearnersEstimatingHeterogeneous2019}.

\textit{Bayesian Additive Regression Trees} (\textbf{BART}) are built along the same principles, using regression trees instead of neural nets; in \cite{atheyRecursivePartitioningHeterogeneous2016}, the BART approach is extending to yield confidence intervals. In \cite{wagerEstimationInferenceHeterogeneous2018}, BART is learned using \textit{Causal Forests} (\textbf{CF}) where the last split of the forest trees corresponds to the treatment assignment.

\subsection{\tlearners\ \label{par:tlearners}} 

T(wo)-learners differ from S-learners in that they learn different models for both potential outcomes. Inspired by domain adaptation \cite{johanssonGeneralizationBoundsRepresentation2022},  \cite{shalitEstimatingIndividualTreatment2017} define a single latent space for both control and treated samples (more in \cite{caronEstimatingIndividualTreatment2022}). On this latent space, two independent functions are learned to estimate the potential outcomes: 
\begin{align*}
    \hmu^0(x) \approx \mu^0(x) = \EE[Y|X=x,T=0] \\
    \hmu^1(x) \approx \mu^1(x) = \EE[Y|X=x,T=1]
\end{align*}
with the individual treatment effect being likewise defined as their difference ($\htau_T(x) = \hmu^1(x) - \hmu^0(x)$). 
In \cite{shalitEstimatingIndividualTreatment2017}, the \textit{BNN} \slearner\ is extended to the \tlearner\ setting.  The embedding $\phi$ is trained to minimize the Wasserstein distance \citep{cuturiSinkhornDistancesLightspeed2013} or the Maximum Mean Discrepancy \citep{grettonKernelTwoSampleTest2012a} between the image of the control and treatment distributions. The impact thereof is inspected by contrasting \textbf{CFR} (constrained distance) and   \textbf{TARNet} (unconstrained). Other approaches aim to preserve local similarity information in the latent space, such as \textbf{SITE} \cite{yaoRepresentationLearningTreatment2018c} or \textbf{ACE} \citep{yaoACEAdaptivelySimilarityPreserved2019}. \textbf{CFR-ISW} \citep{hassanpourCounterFactualRegressionImportance2019} and \textbf{BWCFR} \cite{assaadCounterfactualRepresentationLearning2021a} use \textit{context-aware} weights, reweighting samples in the latent space according to their estimated propensity scores. \textbf{MitNet} \citep{guoEstimatingHeterogeneousTreatment2023b} differs from  \textbf{CFR} by relying on the mutual information between control and treatment distribution, with the advantage that it handles non-binary treatment settings. \textbf{StableCFR} \citep{wuStableEstimationHeterogeneous2023} bridges the gap between \tlearners\ and matching methods \citep{stuartMatchingMethodsCausal2010a}: minority distribution is up-sampled using nearest-neighbor approaches in latent space.

Taking inspiration from adversarial domain adaptation \citep{ganinDomainAdversarialTrainingNeural2016}, \textbf{ABCEI} \citep{duAdversarialBalancingbasedRepresentation2021a} leverages the mutual information between the observed and the latent representation to limit the loss of information; \textbf{CBRE} \citep{zhouCycleBalancedRepresentationLearning2021} considers an auto-encoder architecture with a specific cycle structure: the loss of information due to the latent representation is prevented by requiring this latent information to support the reconstruction of the samples. \\

Generative models are also leveraged for CATE. \textbf{CEVAE} \citep{louizosCausalEffectInference2017c} combines the approach in \cite{shalitEstimatingIndividualTreatment2017} with a Variational Auto-Encoder (\textit{VAE}) \citep{kingmaAutoEncodingVariationalBayes2014d, rezendeStochasticBackpropagationApproximate2014a}; \textbf{GANITE} \citep{yoonGANITEEstimationIndividualized2018} is based on Generative Adversarial Networks (\textit{GAN}) \citep{goodfellowGenerativeAdversarialNets2014}.
\textbf{NSGP} \citep{alaaLimitsEstimatingHeterogeneous2018c} and \textbf{DKLITE} \cite{zhangLearningOverlappingRepresentations2020} are based on Gaussian processes, enabling to minimize counter-factual variance and to provide uncertainty intervals. \\

Refined neural architectures distinguish confounding variables (covariates that cause both $T$ and $Y^0,Y^1$), and adjustment variables (causes of $Y^0,Y^1$ only), in a linear ($\textbf{D}^2\textbf{VD}$ \citep{kuangTreatmentEffectEstimation2017}) or non-linear setting ($\textbf{N-D}^2\textbf{VD}$ \citep{kuangDataDrivenVariableDecomposition2022}). \textbf{DR-CFR} \citep{hassanpourLearningDisentangledRepresentations2019} improves on such refined architectures, by introducing one latent representation for instrumental variables (causes of $T$ only), one for confounding variables, and one for adjustment variables. The latent space is trained in various ways: by minimizing the \textit{MMD} between the latent adjustment representation of control and treated samples \citep{hassanpourLearningDisentangledRepresentations2019}, 
by leveraging Contrastive Log-Ratio Upper Bound \citep{chengCLUBContrastiveLogratio2020} in \textbf{MIM-DRCFR} \citep{chengLearningDisentangledRepresentations2022}; by using a deep orthogonal regularizer in \textbf{DeR-CFR} \citep{wuLearningDecomposedRepresentations2023}; by enforcing the disentanglement using adversarial learning \cite{chauhanAdversarialDeconfoundingIndividualised2023a}. 
The latent structure is combined with a variational approach in \textbf{TEDVAE} \citep{zhangTreatmentEffectEstimation2021}; it is yet further refined in \textbf{SNet} \citep{curthNonparametricEstimationHeterogeneous2021a}, distinguishing adjustment factors causing $Y^0$ only, $Y^1$ only, and both of them. 

\subsection{Other approaches \label{par:otherlearners}} 
\xlearners\ \cite{kunzelMetalearnersEstimatingHeterogeneous2019,stadieEstimatingHeterogeneousTreatment2018a,curthNonparametricEstimationHeterogeneous2021a}
involve a two-step process. In the first step, models of the response functions $\hmu^0,\hmu^1$ and propensity $\heta$ are learned. In the second step, two \CATE\ estimates are trained: $\htau^1(x_i) \approx y_i^1 - \mu^0(x_i)$ is optimized on treated samples, while $\htau^0(x_i) \approx \mu^1(x_i) - y_i^0$ is optimized on control ones. Finally, the \CATE\ estimate for any given sample is obtained as 
\begin{equation*}
    \htau_X(x) = (1-\heta(x))\htau_0(x) + \heta(x)\htau_1(x)
\end{equation*}

\textit{R(obinson)-learners} \cite{nieQuasioracleEstimationHeterogeneous2021} extend the \CATE\ typology defined by \cite{kunzelMetalearnersEstimatingHeterogeneous2019}, building upon the Robinson's potential outcome formalization \citep{robinsonRootNConsistentSemiparametricRegression1988}: 
\begin{equation}
Y - \EE[Y|X] = (T - \EE[T|X])\tau(X) + \varepsilon
\label{eq:robinson}
\end{equation}
with $\varepsilon$ a centered noise variable. Like \xlearners, \rlearners\ proceed along a two-stage
approach: they first learn  $\hat{m}(x) \approx \EE[Y|X=x]$ and $\heta(x) \approx \EE[T|X=x]$, and in a second stage the \CATE\ estimate $\htau_R$ is learned by minimizing: 
\begin{equation*}
    \sum_i \big( y_i - \hat{m}(x_i) - (t_i - \heta(x_i))\htau_R(x_i) \big)^2
\end{equation*}
Using cross-fitting training procedures, this method provably achieves an optimal convergence rate, i.e.,  as if the true $\eta$ and $m$ were known.\\

Related approaches include \flearners\ \citep{kunzelMetalearnersEstimatingHeterogeneous2019} and \ulearners\ \citep{signorovitchIdentifyingInformativeBiological2007, atheyRecursivePartitioningHeterogeneous2016, curthNonparametricEstimationHeterogeneous2021a} using other decompositions of the potential outcome models: \drlearners\ \cite{fosterOrthogonalStatisticalLearning2023, kennedyOptimalDoublyRobust2023} introduce a double robustness methodology \citep{chernozhukovDoubleDebiasedNeyman2017}; \textit{B-learners} \citep{oprescuBlearnerQuasioracleBounds2023} generalize them in settings with a limited amount of unobserved confounders; and \textit{IF-learners} \citep{curthEstimatingStructuralTarget2021}  leverage efficient influence functions \citep{hampelRobustStatisticsApproach1986} to enforce double robustness.

\subsection{Discussion}
The state of the art shows the value of a change of representation for tackling \CATE; \xlearners, unable to align the control and processing distributions, are at a disadvantage.

The search for a latent space faces two 
difficulties.  On one hand, there is  no consensus about how to align both distributions (using ad hoc penalization terms  \citep{johanssonLearningRepresentationsCounterfactual2016a}, \textit{MMD} \cite{shalitEstimatingIndividualTreatment2017}, Wasserstein distance, adversarial learning \cite{duAdversarialBalancingbasedRepresentation2021a}). On the other hand, the latent space is meant to enforce this alignment for both potential outcome models, and must thus achieve some trade-off between both. 

Lastly, \cite{zhangLearningOverlappingRepresentations2020} suggests that the latent space should yield a low \textit{counter-factual variance} (as opposed to, aligning the control and treated distributions). Formally, high variance on the counter-factual posterior distribution suggests that there is not enough information in the considered region regarding the counter-factual modeling task. How to evaluate the counter-factual variance
outside of the Bayesian framework, however, remains an open question.

\section[\XX]{Asymmetrical Latent Regularization for Individual Treatment Effect Modeling \label{chapter:alrite}}

This section first presents the intuition underlying the proposed \XX\ approach. 
After an overview of \XX,
the theoretical analysis of the approach, upper bounding the estimation error under mild assumptions, is described 
and its scope is discussed. 

\subsection{Intuition \label{sub:desiredproperties}}
Following the above discussion, our claim is that CATE must define \textit{two} latent spaces,  allowing counterfactual modeling for both control and treatment samples, i.e. with low counterfactual variance for each distribution.
With no loss of generality let us characterize the counter-factual variance w.r.t. treated samples, with $\phi$ an embedding from input space $\mathcal{X}$ onto latent space $\mathcal{Z}$. 

Given $(x_i,t_i=1,y_i)$ a treated sample, the aim is to estimate $\EE[\YT-\YC|X=x_i]$. Under the assumption that $\phi$ is injective, this estimate coincides with  $\EE[\YT-\YC|\phi(X)=\phi(x_i)]$ (the injectivity requirement is relaxed in section \ref{sec:alrite-discu}).
The variance of the counter-factual estimate for $(x_i,t_i=1,y_i)$ depends on how far $\phi(x_i)$ is from $\phi(x_j)$ for $x_j$ among the control samples. 

As illustrated in \cref{fig:latent}, when a treated point is away from control points in latent space, estimating its counter-factual outcome can be viewed as an out-of-distribution estimation problem. The estimate can be 
 \textit{arbitrarily inaccurate} (unless strong assumptions, e.g., linearity or high smoothness, are made on the potential outcome models). The high uncertainty on the counter-factual estimate is all the greater the higher the dimension of covariate $X$ and the smaller the number $n$ of samples (as is generally the case). This suggests that no treated (respectively, control) sample should be isolated from the control (resp., treated) samples in latent space. 

\begin{figure}
\centering
\begin{subfigure}[ht]{0.8\textwidth}
    \includegraphics[width=1.\textwidth]{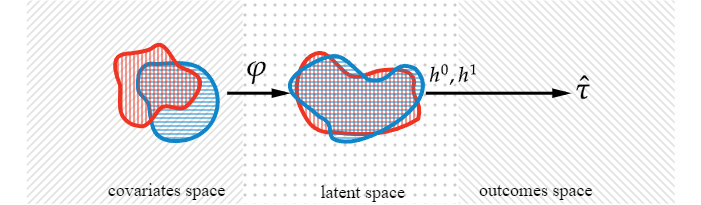}
    \caption{$T$-learner architecture.}
\end{subfigure}
\begin{subfigure}[ht]{0.8\textwidth}
    \includegraphics[width=1.\textwidth]{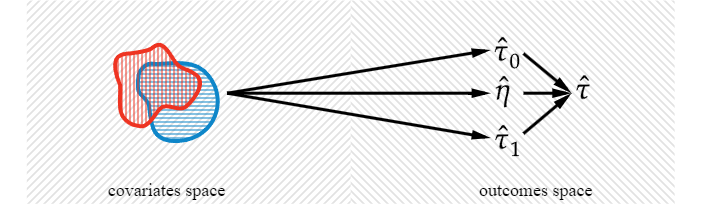}
    \caption{$X$-learner architecture.}
\end{subfigure}
\caption[Schematic comparison of the \textit{T-} and \xlearner\ architectures]{Contrasting the \textit{T-} and \xlearner\ architectures.}
\end{figure}

\def\mtiphi{{x_i^{m,\phi}}}
\def\ytiphi{{y_i^{m,\phi}}}
\def\mti{{x_i^m}}
\def\yti{{y_i^m}}
\def\cfi{\cf$_i$}
\def\iwi{\mbox{$w_i$}}
\def\iwj{\mbox{$w_j$}}
\def\iwk{\mbox{$w_k$}}

\subsection{Overview of \XX \label{sec:principles}}

\subsubsection{Counter-factualizability \label{subsub:counterfactualizability}}
Extending \citep{johanssonLearningRepresentationsCounterfactual2016a, wuStableEstimationHeterogeneous2023}, with $\phi$ an embedding from the covariate space $\mathcal{X}$ onto latent space $\mathcal{Z}$ we define the latent mirror twin of sample $(x_i,t_i,y_i)$ noted 
$(\mtiphi, 1-t_i, \ytiphi)$ as the nearest sample with different treatment assignment in latent space:

\begin{equation}
    (\mtiphi, 1-t_i, \ytiphi) = \argmin \{ \| \phi(x_i) - \phi(x_j) \|, (x_j, t_j = 1 - t_i, y_j) \in {\cal D} \}
    \label{def:mt}
\end{equation}
where the superscript $\phi$ is omitted when clear from the context.\\
The \textit{counter-factualizability} of $x_i$
is defined as its Euclidean distance in latent space to its latent mirror twin ($\|\phi(x_i) - \phi(\mti)\|$): the smaller the better.
As said, $\mti$ is used to estimate the counter-factual outcome for its twin $x_i$. Accordingly, a sample that is mirror twin for several other samples matters more for counter-factual estimation, everything else being equal. This intuition is formalized by defining the \textit{counter-factual importance weight}, noted \iwj, as:
\[ \mbox{\iwj} = \# \{ i \in \onen\ \textit{s.t.}\ (\mtiphi, 1-t_i, \ytiphi) = (x_j, t_j, y_j) \} \]

\begin{figure}
\centering
\begin{subfigure}[t]{.4\textwidth}
  \centering
  \includegraphics[width=1.\linewidth]{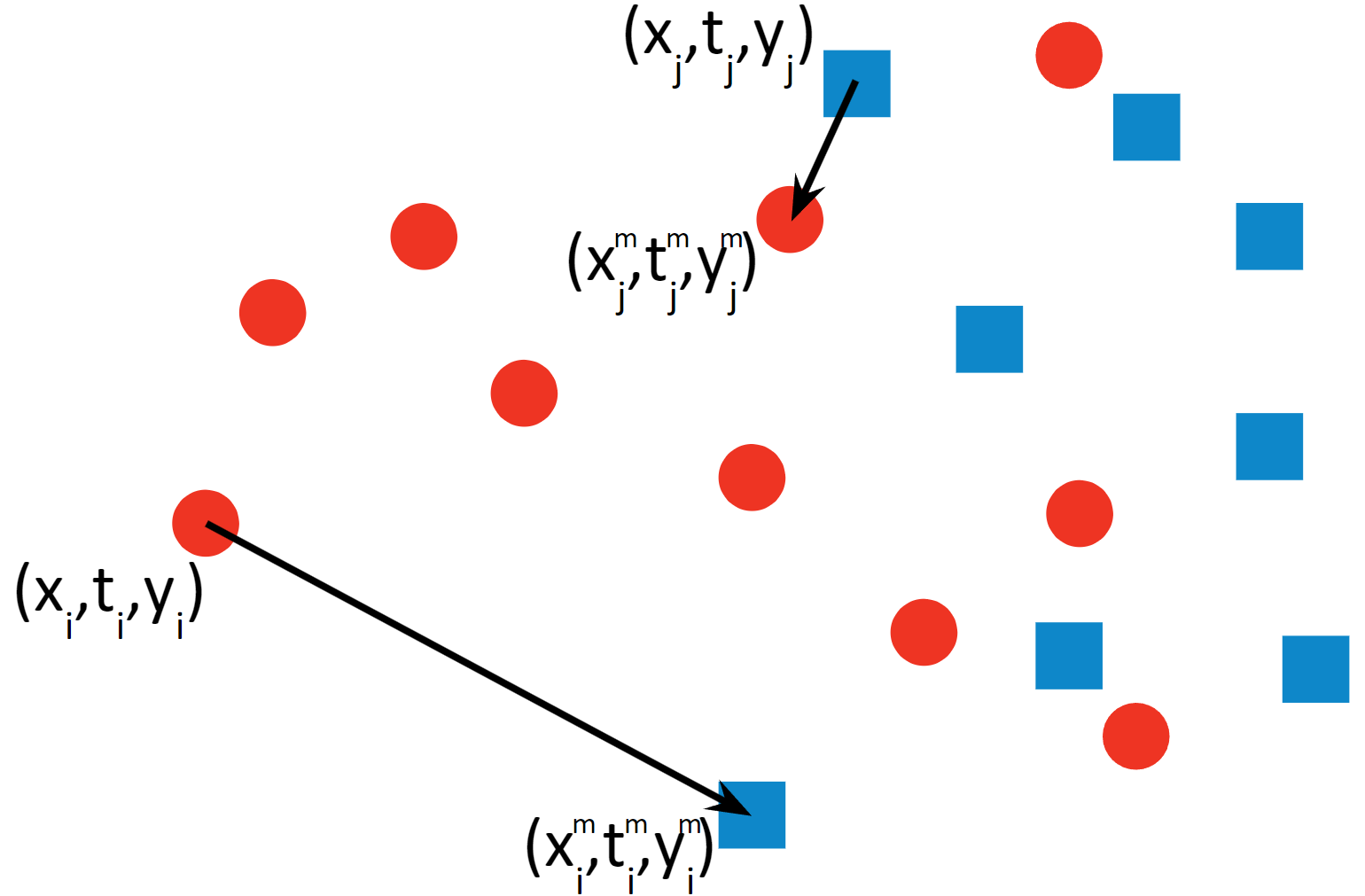}
  \label{fig:mirrortwin}
\end{subfigure}
\hspace{1cm}
\begin{subfigure}[t]{.4\textwidth}
  \centering
  \includegraphics[width=1.\linewidth]{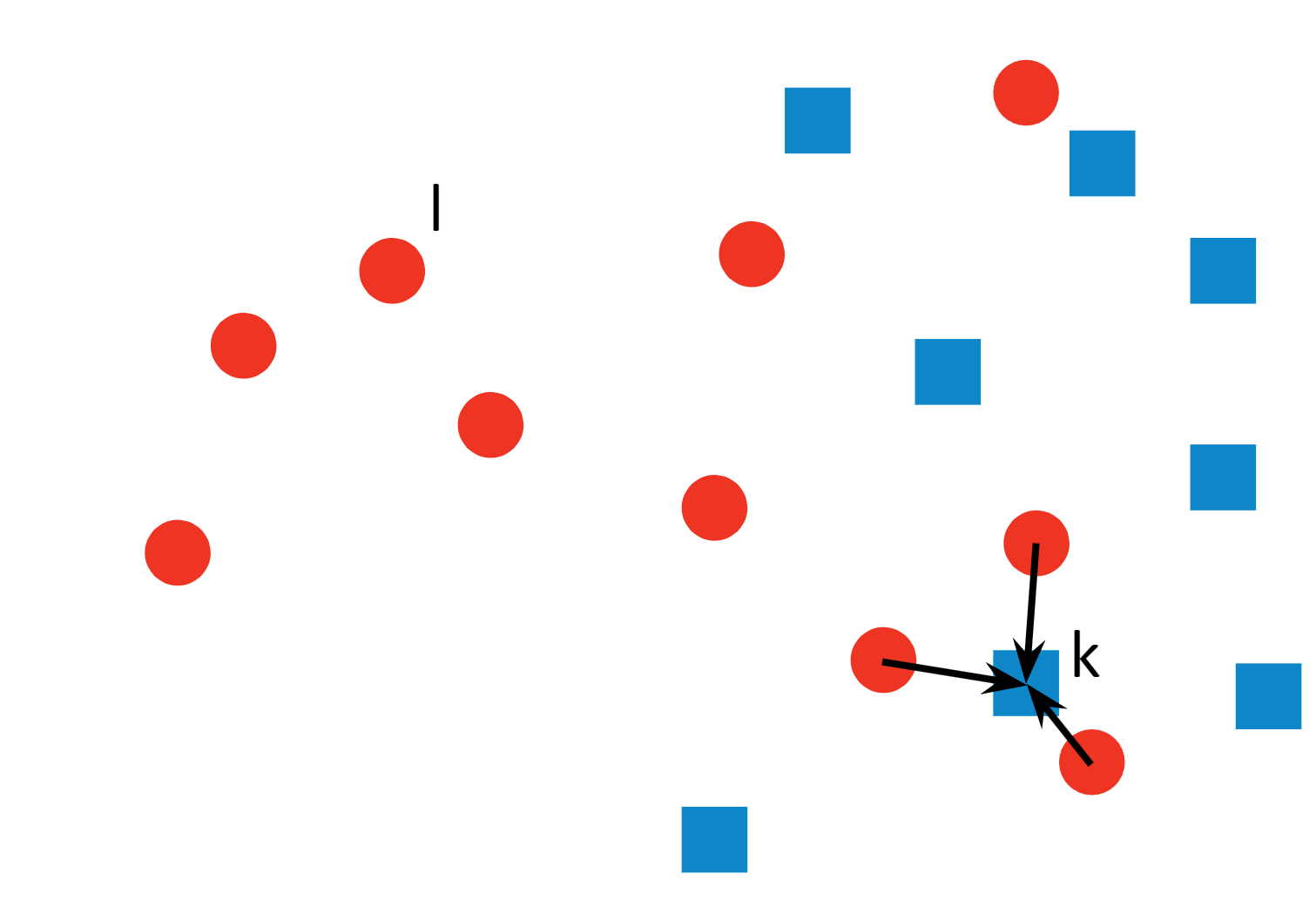}
  \label{fig:isolation}
\end{subfigure}

\caption{\textcolor{C2}{Treated} and \textcolor{B4}{control} samples in latent space. Left: the treated $x_i$ has a far mirror twin; the control $x_j$ has a close mirror twin. Right: the treated $x_\ell$ has counter-factual importance weight 0; the control sample $x_k$ has \iwk=3. \label{fig:latent}}
\end{figure}
The notions of latent mirror twin, and counterfactual importance weight are used to estimate the quality of the latent space. Informally speaking, a latent space allowing good counterfactual estimation of the treated samples is such that: i) the treated examples are close to their mirror twin; 
ii) the $\mu^0(x) \approx \EE[Y | T=0, X=x]$ model learned on this latent space is of good quality, especially for control $x_j$ with a high \iwj\ (since the quality of $\mu_0(x_j)$ has an impact on the counterfactual estimation of many treated $x_i$). 

Formally, a \textbf{pipeline} $\mathcal{P} =(\phi,h^0,h^1)$ is defined as the triplet formed by an embedding $\phi$ mapping the instance space on some latent space ($\phi:\mathcal{X}\rightarrow\mathcal{Z} \subset \RR^d$) and models $h^0$ and $h^1$ of the treated and control outcome ($h^t:\mathcal{Z}\rightarrow\RR$). Note that pipeline $\mathcal{P}$ constitutes a $T$-learner and yields a \CATE\ estimate defined as:
$$\tau(x) = h^1\circ\phi(x) - h^0\circ\phi(x)$$


\subsubsection{Model architecture}
As said (\cref{sub:desiredproperties}), the accurate counter-factual estimate of samples requires these samples to be {\it {counter-factualizable}}, i.e. close to their mirror twins. As depicted on Fig. \ref{fig:2DA} however, the requirements of control and treated samples being counter-factualizable are not necessarily satisfied by the same change of representation.

Consequently, we consider \textbf{two latent spaces}, aimed at ensuring counterfactualization of treated and control samples. More formally, the  \textit{treatment-driven} pipeline \Po\ focuses on CATE for treated samples, while the \textit{control-driven} pipeline \Pz\ focuses on CATE for control samples.

The combination of the CATE estimates built from \Pz\ and \Po\ is classically defined as:
\begin{equation}
    \htau : x\in\mathcal{X} \mapsto (1-\heta(x))\htau_0(x) + \heta(x)\htau_1(x)
\label{eq:tau}
\end{equation}
with $\heta(x)$ estimating the propensity score $P(T=1 | X=x)$ and $\htau_0$ and $\htau_1$ denoting the CATE estimates derived from pipelines \Po\ and \Pz. 

Note that \XX\ bridges the gap between 
$T$-learners and $X$-learners: each pipeline (\Pz\ and \Po) defines a $T$-learner and a causal estimate; these causal estimates are combined as in \xlearners\ (\cref{eq:tau}). 

\begin{figure*}[ht]
\centering
\includegraphics[width=.7\textwidth]{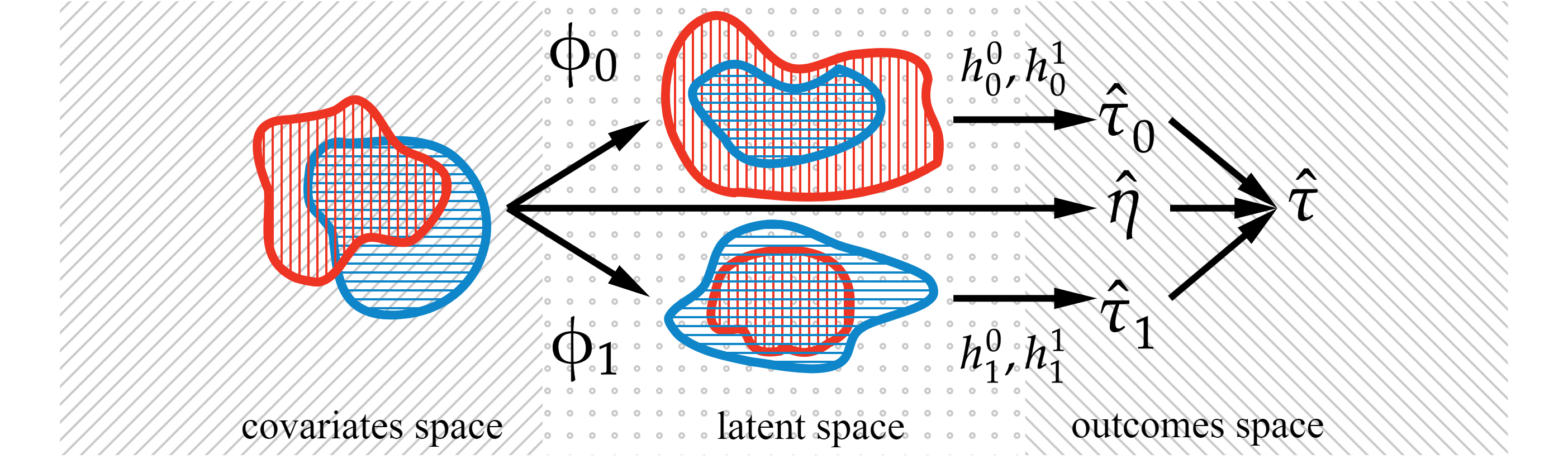}
\caption{\XX\ architecture: Top: control-driven pipeline \Pz. Middle: propensity estimate $\heta$. Bottom: treatment-driven pipeline \Po. Right: overall \CATE\ estimate. }
\label{fig:protocol}
\end{figure*}
\def\hz{h^0}
\def\ho{h^1}
\subsubsection{Training loss}
Let us present the loss used to train control pipeline \Pz = $(\phi, \hz, \ho)$ (the \Po\ loss follows by symmetry). 
\Pz\ is trained end-to-end by minimizing a compound loss enforcing: i) the accuracy\footnote{The accuracy loss depends on the type of the outcome variable $Y$: mean square error is used for continuous $Y$, which is the case considered in the remainder; cross-entropy loss is used for binary $Y$.} of model $\hz$ on the control samples; ii) the accuracy of model $\ho$ on the treated samples (accounting for the fact that control samples with high counterfactual importance weight matter more); iii) the good counter-factualizability of control samples (a small distance to their mirror twin); iv) the regularization of the whole pipeline vector weight noted $\|\Pz\|$:
\begin{align}
\begin{split}
    {\cal L}(\Pz) = &\frac{1}{n_0}\sum_{t_i=0} (y_i - \hz(\phi(x_i)))^2 + \frac{1}{n_1+\beta n_0} \sum_{t_i=1} (1 + \beta \iwi) (y_i - \ho(\phi(x_i)))^2 \\
    & + \alpha \frac{1}{n_0}\sum_{t_i=0} \|\phi(x_i) - \phi(\mti) \|^2 + \gamma \|\Pz\|^2
\end{split}
\label{eq:loss}
\end{align}
with $n_0$ and $n_1$ the number of control and treated samples in $\cal D$, $\iwi$ the counterfactual importance weight of $x_i$, and $\alpha, \beta, \gamma$  hyper-parameters of the approach.\footnote{In all generality, the hyper-parameters used to train \Pz\ and \Po\ do not need to be the same. The hyper-parameter setting is detailed in \cref{chapter:hyperparameter}.}

\subsection{Algorithm \label{sec:algo}}
The \XX\ algorithm finally involves three modules: learning \Pz\ (Eq. \ref{eq:loss}); learning \Po; learning the propensity estimate $\heta$. Finally, the \CATE\ estimates derived from \Pz\ and \Po\ are aggregated using $\heta$ (Eq. \ref{eq:tau}). 

\def\hzz{h_{0}^{0}}
\def\hzo{h_{0}^{1}}
\def\hoz{h_{1}^{0}}
\def\hoo{h_{1}^{1}}

\begin{algorithm}
\caption{\XX \label{alg:xx}}
\begin{algorithmic}
\Require Dataset $\mathcal{D}$; hyper-parameters $\alpha, \beta, \gamma$.  
    \State \textbf{Learn} \Pz =  $(\phi_0, \hzz, \hzo)$ (Eq. \ref{eq:loss})
    \State Compute $\htau_0$
    \State \textbf{Learn} \Po = $(\phi_1, \hoz, \hoo)$ 
    \State Compute $\htau_1$
    \State \textbf{Learn} $\heta$  // \textit{Propensity score}
    \State Compute $\htau$ (Eq. \ref{eq:tau})
\end{algorithmic}
\end{algorithm}

As said, the propensity score \heta\ can be learned using classical supervised learning. Formally, a cross-entropy loss is used:
\begin{equation}
    \mathcal{L}(\heta) = -\frac{1}{n_1}\sum_{t_i=1} \log\big(\heta(x_i)\big)-\frac{1}{n_0}\sum_{t_i=0}\log\big(1-\heta(x_i)\big) + \|\heta \|^2
\end{equation}
with $\|\heta\|^2$ a regularization term. 
It is desirable that $\heta$ be calibrated \citep{zadroznyTransformingClassifierScores2002}, i.e. such that 
$$\forall s\in [0,1],\ \EE[T=1|\heta(X)=s]=s$$ 

However the sensitivity of the overall $\htau$ (defined by aggregating the \CATE\ estimates $\htau_0$ and $\htau_1$ respectively derived from \Pz\ and \Po, Eq. \ref{eq:tau}) w.r.t. the propensity score is limited. Let $\tau_{\eta}$ denote the aggregation of $\htau_0$ and $\htau_1$ using the ground truth $\eta$. It reads: 
\begin{equation}
\begin{aligned}
    ||\htau - \htau_{\eta}|| &= ||(\heta-\eta)(\htau_0-\htau_1)|| \\
    & \le ||\heta-\eta|| \times \big(\|\htau_0-\tau\| + \|\htau_1-\tau \|\big)
\end{aligned} 
\label{eq:etaerror}
\end{equation}
In other words, the \CATE\ error due to the error on $\eta$ is of order 2, as the product of i) the error on the propensity; ii) the error on $\htau_0$ and on $\htau_1$. 

{\em Remark}: It is straightforward to define an ensemble variant of \XX, by splitting the dataset into a training, a validation, and a test dataset. On the training subset, several pipelines \Pz\ and \Po\ are learned using different hyper-parameter settings. The best pipelines (determined from their factual accuracy on the validation set) are selected and they are aggregated to form the ensemble \CATE\ model. The aggregation is the average of the top-K out of all pipelines, with $K$ a hyper-parameter of the ensemble approach, or a weighted sum, where the weight of each model corresponds to a softmax of hyper-parameter $\lambda$ (Appendix \ref{sec:ensemble}).

\subsection{Formal guarantees \label{sec:analysis}}
As said, a main contribution of the approach is to provide formal guarantees on the \PEHE\ error of \XX, considering the within-sample setting, i.e. when the treatment variable $T$ is known. Remind that the within-sample error is not trivial $-$ contrarily to the training error in supervised learning $-$ as counter-factual $Y^{1 - T}$ is not observed.
All proofs are given in Appendix 
\ref{appendix:proofs}.
Our first result states that the \PEHE\ loss associated with a \tlearner\ is upper bounded depending on the counter-factualizability of the samples. 
\begin{theorem} \label{theorem}
Let $\phi: \mathcal{X} \rightarrow \mathcal{Z}$ be an embedding from $\mathcal{X}$ to a latent space $\mathcal{Z}$. Let us assume that the sought outcome models $\mu^0$ and $\mu^1$ can be expressed as $\mu^0 = \nu^0 \circ \phi$ and $\mu^1 = \nu^1 \circ \phi$, with $\nu^0$ and $\nu^1$ two functions defined on $\mathcal{Z}$ with Lipschitz constant $L$. \\
Let $\hnu^0, \hnu^1 : \mathcal{Z} \rightarrow \RR$ be two models trained to approximate $\nu^0$ and $\nu^1$ with Lipschitz constant $\hat{L}$. 
Then the empirical $\PEHE$ associated with $(\hnu^1\ - \hnu^0) \circ \phi$ is upper bounded by $M_1$ with: 
\begin{equation*}
    M_1 =\frac{4}{n} \Big[
    \sum_{i=1}^{n} \Big(1+ \iwi\Big)
    \Big((\hnu^{t_i} \circ \phi) (x_i) - y_i \Big)^2
    + (L^2 + \hL^2) \sum_{i=1}^{n} \| x_i - \mti\|^2 \Big]
\end{equation*}
\end{theorem}
Bound $M_1$ depends on two terms: i) the factual accuracy on every sample $x_i$, all the more important the higher its counter-factual importance weight \iwi, and ii) the counter-factualizability of the samples. This bound establishes the soundness of the training loss (Eq. \ref{eq:loss}), built on both terms.

Building on this result, our second result concerns the hybrid \xlearner\tlearner\ scheme of \XX. 

\def\nuzz{\nu_{0}^{0}}
\def\nuzo{\nu_{0}^{1}}
\def\nuoz{\nu_{1}^{0}}
\def\nuoo{\nu_{1}^{1}}
\begin{theorem}
Let $\Pz$ and $\Po$ be a control and a treatment pipeline, respectively involving embeddings $\phi_0$ and $\phi_1$. Let us further assume that the sought outcome models $\mu^0$ and $\mu^1$ can be expressed on the top of each embedding ($\mu^0 = \nuzz \circ \phi_0 = \nuoz \circ \phi_1$ and $\mu^1 = \nuoz \circ \phi_0 = \nuoo \circ \phi_1$), with all $\nu_{i}^{j}$ of Lipschitz constant $L$ for $i,j \in \{0,1\}$. \\
Let $\hzz$ and $\hzo$ (respectively $\hoz$ and $\hoo$) denote the learned models of 
$\nuzz$ and $\nuzo$ in \Pz\ (resp. $\nuoz$ and $\nuoo$ in \Po), with Lipschitz constant $\hL$. 

For any 
$i$-th sample in the observation dataset $\cal D$, define 
$$\Bar{\tau}_i = (1-t_i)\big((\hzo\circ\phi_0)(x_i)-y_i\big) + t_i \big(y_i - (\hoz\circ\phi_1)(x_i)\big)$$
Let us further define the counter-factual importance weight of sample $(x_j,t_j,y_j)$ as the number of samples $(x_i,t_i,y_i)$ such that $t_i=1-t_j$ and $x_i$ is the nearest neighbor of $x_j$ according to $\phi_{t_i}$:
$$ \iwj = \#\{ (x_i, t_i, y_j) \in \DD\ s.t.\ (x_j, t_j, y_j) = (x_i^{m,\phi_{t_i}}, 1 - t_i, y_i^{m, \phi_{t_i}}) \}$$ 

Then, the within-sample \PEHE\ defined by $\frac{1}{n}\sum_{i=1}^n \big(\Bar{\tau}_i -\tau(x_i)\big)^2$ is upper bounded by $M_2$, with 
\begin{equation*}
    M_2 = \frac{5}{n} \Big[
    \sum_{t_i = 1, i=1}^n \iwi \bigl((\hzo \circ \phi_0)(x_i) - y_i\bigr)^2 
    + \sum_{t_i = 0, i=1}^n \iwi ((\hoz \circ \phi_1)(x_i) - y_i)^2 
    + (L^2+\hL^2) \sum_{i=1}^n \| \phi_{t_i}(x_i) - \phi_{t_i}(x_i^{m,\phi_{t_i}})\|^2  + \kappa_Y \Big]
\end{equation*}
where $\kappa_Y = \sum_{i=1}^n \big(1+\iwi\big)\big(y_i-\mu^{t_i}(x_i)\big)^2$.
\label{theorem2}
\end{theorem}

As said, \cref{theorem2} holds in the \textit{within-sample} setting, as it requires knowledge of the treatment assignment $T$. It does not generalize directly to the \textit{out-of-sample} setting, where the (unknown) $t_i$ and $y_i$ are respectively estimated using the propensity and the outcome models. \\

Finally, the upper-bound established in \cref{theorem2} is directly related with the loss used to train the pipelines (Eq. \ref{eq:loss}), establishing the well-foundedness of the approach, as follows: 

\begin{theorem}
There exists a hyper-parameter setting $(\alphaz,\alphao,\betaz,\betao)$ such that the \textit{within-sample} empirical \PEHE\ is upper bounded by $M_3$, with:
\begin{align*}
    M_3 =& 5 \Big[ {\cal L}(\Po)  + {\cal L}(\Pz) - \gamma_0 \|\Pz\| - \gamma \|\Po\|  + \kappa_Y\\
    &- \frac{1}{n_0}\sum_{t_i= 0, i=1}^n (\hzz \circ\phi_0 (x_i)-y_i)^2 
    - \frac{1}{n_1}\sum_{t_i= 1, i=1}^n (\hoo \circ\phi_1 (x_i)-y_i)^2\\
    & 
    - \frac{1}{n}\big(\sum_{t_i=0,i=1}^n (\hoz \circ\phi_1(x_i)-y_i)^2 
    + \sum_{t_i=1,i=1}^n (\hzo \circ\phi_0(x_i)-y_i)^2 \big) \Big]
\end{align*}
\label{theorem3}
\end{theorem}
Note that this result is not constructive in the sense that it does not give the hyper-parameter setting. Nevertheless, the relation between this bound on the PEHE and the terms in the pipeline loss confirms the soundness of the approach. This bound also depends on the Lipschitz constants
of the learned models\footnote{In the case where \HCF,\HCCF,\HTCF,\HTF are linear, their Lipschitz constants can be derived straightforwardly. In the general case of neural networks, additional care is required (see e.g., \cite{virmauxLipschitzRegularityDeep2018, goukRegularisationNeuralNetworks2021}).} and of the target models (which only depend on the problem).  


\subsection{Discussion}\label{sec:alrite-discu}
The formal guarantees established by Thm 1-3 contrast with the main result of Shalit et al. \cite{shalitEstimatingIndividualTreatment2017} in two ways.\footnote{
Let us remind this result 
for the sake of self-containedness: 
\begin{theorem*}[from \cite{shalitEstimatingIndividualTreatment2017}]
    Let $(\phi,h^0,h^1)$ be a pipeline such that $\phi:\mathcal{X}\rightarrow\mathcal{Z}$ is invertible.
    Define the point-wise loss function $\ell_{h,\phi}$ and expected factual loss $\epsilon_{t}$ conditionally to treatment assignment $T=t,\ t\in\zeroone$ by
    \begin{equation}
    \begin{cases}
        \ell_{h,\phi} : (x,t)\in\mathcal{X}\times\zeroone &\mapsto \EE[\big(Y^t-h\circ\phi(x)\big)^2|X=x] \\
        \epsilon^t : h\in(\mathcal{Z}\rightarrow\mathcal{Y}) &\mapsto \EE[\ell_{h,\phi}(X,T)|T=t]
    \end{cases}
    \label{eq:shalitnotations}
    \end{equation}
    Denote by $\sigma_{Y^t}^2 = \EE\big[(Y^t-\EE[Y^t|X])^2|T=t\big]\PP(T=t)$ the expected variance of $Y^t$, $t\in\zeroone$.
    Let G be a family of functions $\mathcal{Z}\rightarrow\mathcal{Y}$. Assume that there exists a constant $B_\phi>0$ such that $z\in\mathcal{Z}\mapsto\frac{1}{B_\phi}\ell_{h^t,\phi}(\phi^{-1}(z),t) \in G$, $t\in\zeroone$.
    Denote the integral probability metric between the control and treated latent distributions induced by $\phi$ as
    $$\textit{IPM} = \sup_{g\in G}\big|\EE\big[\big(\PP\big(\phi(X)|T\text{=}1\big)-\PP\big(\phi(X)|T\text{=}0\big)\big)g\big(\phi(X)\big)\big]\big|$$
    Then, the $\PEHE$ is upper bounded:
    $$\PEHE(\phi,h^0,h^1) \leq 2\big(\epsilon^0(h^0)+\epsilon^1(h^1)+B_\phi\times\textit{IPM} -2\min(\sigma_{Y^0}^2,\sigma_{Y^1}^2) \big)$$
\end{theorem*}
}
A first difference is that \cite{shalitEstimatingIndividualTreatment2017} relies on the invertibility of embedding $\phi$. In contrast, \XX\ only assumes that the considered embeddings $\phi$ are such that the sought models can be expressed with no loss of information (there exists $\nu^0$ and $\nu^1$ s.t. $\mu^\ell = \nu^\ell \circ\phi$). As a result, \XX\ can fully benefit of the celebrated opportunities offered by a change of representation \citep{caytonAlgorithmsManifoldLearning2008, bengioRepresentationLearningReview2013}, e.g. to achieve feature selection or dimensionality reduction. 

A second difference is that the bound in \cite{shalitEstimatingIndividualTreatment2017} involves \textbf{integral probability metrics}, thus defined in the large sample limit case. In contrast, our results build upon the counter-factual importance weights and the counter-factualizability of the samples, that is, empirical quantities that are in principle more readily understood by a practitioner. \\


Another key issue concerns the efficiency of the loss, and whether the considered optimization problem admits the sought solution as optimum. Let us focus on the control model; the treatment model case follows by symmetry.
Under the assumption that embedding $\phi$ is such that it entails no loss of information w.r.t. the sought models, i.e. there exists $\nu^t$ such that $\mu^t = \nu^t \circ \phi$, then the ground truth $\nu^t$ minimizes the factual error loss.

\begin{lemma}
    Under the assumption of conditional exchangeability, let embedding $\phi$ be such that there exists $\nu^0$ with $\mu^0=\nu^0\circ\phi$. For any candidate model $\hnu^0$, let ${\cal L}(\hnu^0)$ denote the expected mean squared prediction error of $\hnu^0$ conditionally to $T=0$ :
    \begin{align*}
        {\cal L}(\hnu^0) &= \EE_{X,Y|T=0}\big[\big(\hnu^0\circ\phi(X)-Y^0\big)^2|T=0\big] \\
        &= \EE_{Z=\phi(X),Y|T=0}\big[\big(\hnu^0(Z)-Y\big)^2|T=0\big]
    \end{align*}
    Then, $\nu^0$ reaches the minimum of $\cal L$.
\label{lemma:minimizing}
\end{lemma}
Proof: in Appendix 
\ref{appendix:proofs}.
Note that from the practitioner's viewpoint, it is impossible to prove that a given statistic is sufficient based only on observational data. Indeed if $\phi$ is injective then $Y^0\indep X|\phi(X)$, but this result relies on mapping $\phi$, and not on the observational data itself. Even in the large sample limit and using adequate conditional independence statistical tests, one may prove at most independence of $Y^0$ and $X$ conditionally to $(\phi(X)=z,T=0)$, but not conditionally to $(\phi(X)=z)$ alone. This concern echoes the ones raised by the assumption of conditional exchangeability: assuming sufficiency based on observational data is a similar leap of faith. \\

\begin{figure}
    \begin{subfigure}{1.\textwidth}
    \centering
    \includegraphics[width=.4\textwidth]{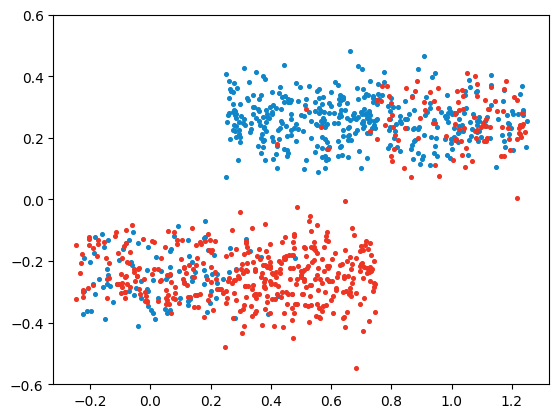}
    \label{fig:toyproblempositivity}
    \end{subfigure}
    \begin{subfigure}{.48\textwidth}
      \centering
      \includegraphics[width=.7\linewidth]{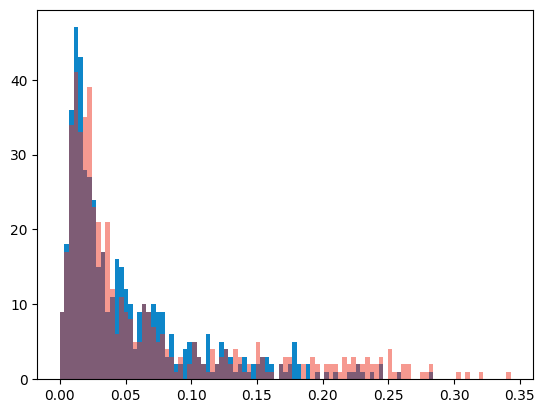}
    \end{subfigure}
    \hfill
    \begin{subfigure}{.48\textwidth}
      \centering
      \includegraphics[width=.7\linewidth]{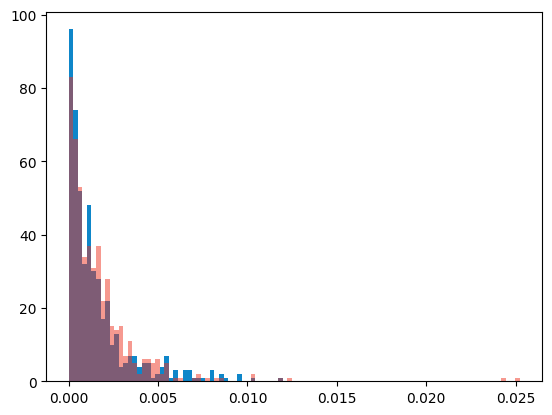}
    \end{subfigure}
\caption{Distribution of the \textcolor{B4}{control} and \textcolor{C2}{treated} samples. Top: covariate space $\RR^2$. Bottom left: distribution of distances to mirror twins for $\phi = $Id. Bottom right:  distribution of distances to mirror twins for $\phi = $projection on the $x$-axis.} 
\label{fig:fxxxdisplay}
\end{figure}

Let us last discuss the limitations of the approach and its robustness. 
The synthetic data depicted on Fig. \ref{fig:fxxxdisplay} illustrates the sensitivity of \XX\ w.r.t. the violation of the positivity assumption: the rightmost samples of the bottom cluster are overwhelmingly \textcolor{C2}{treated}, and the leftmost samples of the top one are mainly \textcolor{B4}{control}. The average distance to the mirror twin is much better for $\phi$ set to the projection on the $x$-axis (Fig. \ref{fig:fxxxdisplay}, bottom right) than for $\phi = $Id (Fig. \ref{fig:fxxxdisplay}, bottom left). The former $\phi$ might thus correspond to a local minimum of the training loss due to the low counter-factualizability, at the expense of the factual accuracy of the learned models.\footnote{It is fair to say however that approaches based on the minimization of distributional distance \citep{shalitEstimatingIndividualTreatment2017, duAdversarialBalancingbasedRepresentation2021a}, or counter-factual variance \citep{zhangLearningOverlappingRepresentations2020} present the same weakness. The weakness is also neither addressed by latent space disentanglement nor double-robustness.} 

Another potential weakness of the approach is due to the impact of samples with high counter-factual importance weight. The compound loss (Eq. \ref{eq:loss}) focuses on optimizing the factual accuracy for samples with a high weight. A risk is to distort the optimization of $\phi$, in such a way that samples that are hard to model have a low counter-factual importance weight. This risk is mitigated as the gradient does not flow back through the mirror twin operator during the back-propagation phase; there should be no incentive for the optimization process to distort $\phi$ in such a pathological way. Further research is concerned with defining a smoother counter-factual importance weight.

\section{Experimental validation}\label{sec:resu}
After describing the two benchmarks we considered, \IHDP\  and \ACIC, and the experimental setting, this section reports on the empirical validation of \XX.
The \XX\ code is publicly available.\footnote{\href{https://github.com/ArmandLacombe/alrite}{Code repository}.
Following \cite{yaoRepresentationLearningTreatment2018c, duAdversarialBalancingbasedRepresentation2021a, zhouCycleBalancedRepresentationLearning2021}, the implementation builds on the code released by \cite{shalitEstimatingIndividualTreatment2017} and \cite{johanssonCfrnet2023}, using \textit{Tensorflow} \citep{abadiTensorFlowSystemLargescale2016} for the training of neural models and  \textit{Scikit-Learn} \citep{pedregosaScikitlearnMachineLearning2011a} for auxiliary models.\\
The main results on \IHDP\ as reported in \cref{tab:resultsIHDP} (resp. \ACIC, as reported in \cref{tab:resultsacic}) have been obtained after randomly selecting $70$ (resp. $16$) sets of hyper-parameters for each pipeline, totaling the training of $14000$ (resp. $12320$) models. The training of each model requires $4$ Intel® Xeon® Silver 4108 
\textit{CPUs} and $2$ (resp. $6$) \textit{GB} of \textit{RAM} on average and a total training (wall) time of $81$s (resp. $343$s).
}

\subsection{Benchmarks \label{sec:benchmarks}}
The \IHDP\ dataset is the baseline dataset commonly used for benchmarking causal inference models \citep{shalitEstimatingIndividualTreatment2017, duAdversarialBalancingbasedRepresentation2021a, zhouCycleBalancedRepresentationLearning2021}. The \ACIC\ benchmark originates from the 2016 \textit{Atlantic Causal Inference Challenge} \citep{dorieAutomatedDoItYourselfMethods2017}.

\subsubsection{\IHDP \label{par:ihdp}}
The \IHDP\ dataset, introduced by \cite{hillBayesianNonparametricModeling2011a}, is based on a real-life randomized experiment dataset, the Infant Health and Development Program \citep{brooks-gunnEffectsEarlyIntervention1992}. The treatment consists of quality child care and home visits on the health and development of preterm children. While the treatment assignment is randomized in the collected data, treatment imbalance is enforced by \cite{hillBayesianNonparametricModeling2011a}, by {removing a subpopulation} (treated group children with non-white mothers) from the dataset. 

The dataset contains $747$ individuals, $139$ of whom are treated; these are described by 25 covariate features, among which 6 are continuous and 19 binary. 

The major difference between the original survey results and \cite{hillBayesianNonparametricModeling2011a}'s \IHDP\ dataset lies in the replacement of the initial outcome with {simulated outcomes}, making it possible to access counter-factual quantities and ensuring that {conditional exchangeability holds}. The selected simulation method consists in defining two response surfaces $\mu^t:x\in\mathcal{X}\mapsto\EE[Y^t|X=x],t\in\zeroone$, to which Gaussian noise is added. 

We consider the \IHDP-100 benchmark, involving 100 datasets (referred to as problem instance or instance when no confusion is to fear), where each dataset is generated by: i) drawing $\beta$ in $[0,1]^{25}$; ii) setting the surface responses as: 
\begin{align*}
    \YC &\sim \exp\{\langle X+0.5, \beta\rangle\}+\mathcal{N}(0,1)\\
    \YT &\sim \langle X+0.5, \beta\rangle + \omega + \mathcal{N}(0,1), \mbox{ with $\omega$ s.t. average treatment effect }\ATT=4
\end{align*}
Each dataset is split into a training (90\% of samples) and a testing (10\%) subset, the split being fixed to enable a fair comparison among the algorithms. Following the state of the art, 30\% of the training set is held out to form a validation set.

The main performance indicators are the {within-sample} and {out-of-sample} \PEHE, respectively computed on the training samples (where the treatment assignment and factual outcome are known, and the counter-factual outcome is unknown) and on the test set (where the treatment assignment and both outcomes are unknown). 
Following the state-of-the-art \citep{shalitEstimatingIndividualTreatment2017, duAdversarialBalancingbasedRepresentation2021a, zhouCycleBalancedRepresentationLearning2021}, the results are averaged over the 100 datasets. 

A secondary performance indicator on the \IHDP\ benchmark is the {mean absolute error on the \ATE\ estimation} $\epsilon_{\textit{ATE}}$ \pcref{eq:metricsdef}. Note that \ATE\ is notorious for being subject to systemic estimation bias: $\epsilon_{\textit{ATE}}$ favors low-bias models (as opposed to \PEHE, that favors low-variance models).

\begin{figure}
\begin{subfigure}[t]{.3\textwidth}
  \centering
  \includegraphics[width=1.\linewidth]{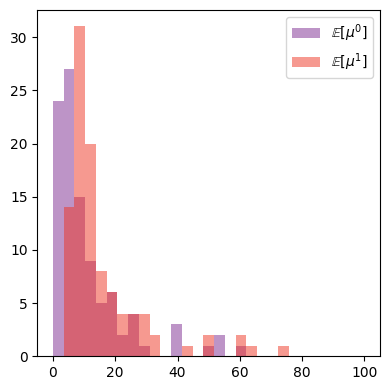}
  \caption{Average $\mu^0$ and $\mu^1$}
  \label{fig:measuresIHDP100mean}
\end{subfigure}
\hfill
\begin{subfigure}[t]{.3\textwidth}
  \centering
  \includegraphics[width=1.\linewidth]{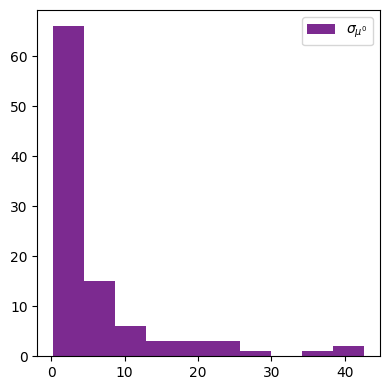}
  \caption{Standard deviation of $\mu^0$}
  \label{fig:measuresIHDP100std0}
\end{subfigure}
\hfill
\begin{subfigure}[t]{.3\textwidth}
  \centering
  \includegraphics[width=1.\linewidth]{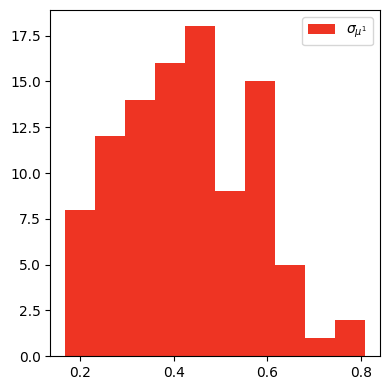}
  \caption{Standard deviation of $\mu^1$}
  \label{fig:measuresIHDP100std1}
\end{subfigure}
\begin{subfigure}[t]{.3\textwidth}
  \centering
  \includegraphics[width=1.\linewidth]{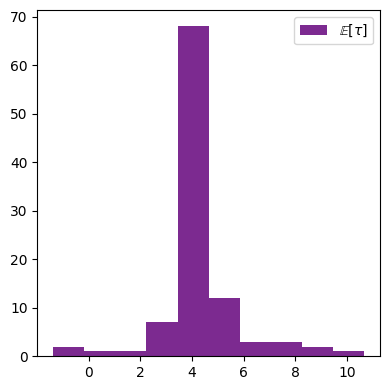}
  \caption{Average treatment effect (ATT).}
  \label{fig:measuresIHDP100ATE}
\end{subfigure}
\hfill
\begin{subfigure}[t]{.3\textwidth}
  \centering
  \includegraphics[width=1.\linewidth]{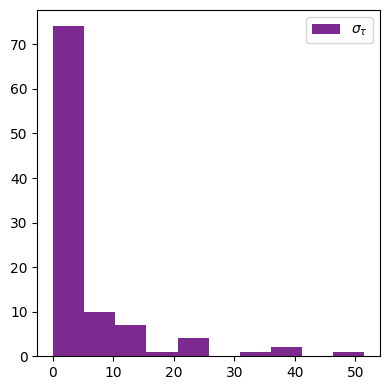}
  \caption{Standard deviation $\sigma_\tau$ of ATT.}
  \label{fig:measuresIHDP100stdTE}
\end{subfigure}
\hfill
\begin{subfigure}[t]{.3\textwidth}
  \centering
  \includegraphics[width=1.\linewidth]{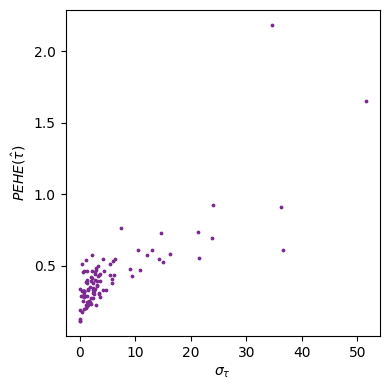}
  \caption{$\PEHE(\htau)$ vs $\sigma_\tau$}
  \label{fig:measuresIHDP100PEHE}
\end{subfigure}
\caption{Diversity of the 100 problem instances in \IHDP-100: Histograms of effects and standard deviations, and relation between error \PEHE\ and standard deviation $\sigma_\tau$.}
\label{fig:IHDPfeatures}
\end{figure}

Some criticisms have been presented in the literature concerning \IHDP\ \citep{curthReallyDoingGreat2021}:

Firstly, the ranges of outcomes and the causal effects are {not commensurate among the different problem instances}, as shown in \cref{fig:IHDPfeatures}(a). Likewise, the standard deviation of $\mu^0$ takes high values in some instances while it is generally low for $\mu^1$. As a result, $\tau$ widely varies depending on the problem instance, both in terms of average and variance.

A high standard deviation $\sigma_{\tau}$ of the causal effects makes \CATE\ estimation more difficult: when $\sigma_{\tau}$ goes to $0$, the causal effect is uniform and the \CATE\ estimation problem boils down to the (much simpler) \ATE\ estimation problem. Quite the contrary, a large $\sigma_{\tau}$ makes large errors more likely. This claim is visually confirmed in \cref{fig:IHDPfeatures}-(f), plotting the standard deviations of $\tau$ vs the \PEHE\ of the \XX\ estimate $\htau$ for all instances in \IHDP-100. Interestingly, instances with large $\sigma_{\tau}$ account for a large fraction of the overall \PEHE\ error. Accordingly, \IHDP\ performance indicators strongly depend on how the algorithm behave on the few toughest instances; they do not reflect the average behavior of the algorithm. 

\subsubsection{\ACIC \label{par:acic}}

\ACIC, designed for  the 2016 \textit{Atlantic Causal Inference Challenge} \citep{dorieAutomatedDoItYourselfMethods2017}, aims to compare different causal inference protocols. Like \IHDP, \ACIC\ is a set of problem instances, generated as follows: $77$ data generation protocols are first defined. For each protocol, in order to limit the computational burden and in accordance with \cite{zhangTreatmentEffectEstimation2021}, we took into account the first $10$ generated datasets, i.e. a total of $770$ datasets.

Each dataset contains $4802$ samples, described by $58$ features derived from real-world data. \cite{lightCollaborativePerinatalStudy1973} presents a longitudinal study carried out between years $1959$ and $1965$, with a view to identifying factors that affect the probability of children's organic neurological defects. $23$ of the $58$ features are continuous, $27$ are counts, $5$ are binary and $3$ are categorical. 

Like for \IHDP, the treatment assignment and potential outcomes are simulated, ensuring that the conditional exchangeability, positivity, and SUTVA assumptions hold true.  Data generation protocols \cite{dorieAutomatedDoItYourselfMethods2017} differ in their degree of nonlinearity, percentage of treatment assignment, overlap, alignment, treatment effect magnitude and heterogeneity, allowing a wide variety of situations to be described.

Formally the data generation process is of the form:
$$
f : x\in\mathcal{X} \mapsto g\big(f_1(x_1)+f_2(x_2)+f_3(x_1)f_4(x_2)\big)
$$
where the $f_i$ functions are either polynomial terms, step functions or indicator functions; the different features may be combined in various ways. The link function  $g$ introduces a nonlinearity or bound the output. The  outcome functions are defined as $Y^t \sim g_t(X) + \mathcal{N}$ with $\mathcal{N}$ a centered noise term; the propensity is defined as: $T \sim \textit{Ber}(f(X))$.


The performance indicator is \PEHE, averaged over all $770$ instances of the \ACIC\ dataset.\footnote{Although the original performance metric of the  \textit{Atlantic Causal Inference Challenge} is the Sample Average Treatment Effect, the treatment effect is by design heterogeneous, making \ACIC\  well-suited to CATE assessment.}

As said, the \ACIC\ benchmark  aims to overcome the limitations of common benchmark datasets \citep{dorieAutomatedDoItYourselfMethods2017}, while enabling to compute the \PEHE. In particular, the choice to extract the covariates from real-life data avoids the potential data homogeneity of synthetic datasets. 
Despite the variety of the potential outcome models, the \CATE\ is the difference between the two surface responses ($\tau = \mu^1 - \mu^0$), thus favoring \tlearners\ and \slearners. \footnote{There exists however another
potential outcome formalization due to \cite{robinsonRootNConsistentSemiparametricRegression1988, nieQuasioracleEstimationHeterogeneous2021} and occasionally referred to as "Robinson's transformation", reading $Y \sim m(X) + (T-\eta(X)) \tau(X) + \mathcal{N}$, where $\eta$ stands for the propensity score and $m$ the conditional average outcome. As discussed in \cite{curthReallyDoingGreat2021}, this other formalization is more favorable to \rlearners.}

\subsection{Baselines \label{sec:expsetting}}

The baseline models considered to comparatively assess the performance of \XX\ are listed in 
\cref{tab:resultsIHDP}.
It is emphasized that their results are those reported in the cited papers: In some cases, the code is not made public and there is a lack of details regarding the implementation; in other cases, there is a lack of details regarding the hyper-parameter setting; both prevent us from reproducing the results.

\subsection{\XX: Hyper-parameter setting}
The detail of the hyper-parameters and computational framework is given in Appendix 
\ref{chapter:hyperparameter}.
Pipelines \Pz\ and \Po\ are trained independently, and implemented as neural networks with  Exponential Linear Units (\textit{ELU})  activation functions \citep{clevertFastAccurateDeep2015}.

The architecture hyper-parameters include the number $L$ of layers and the layer width $W$ (same for all layers), with normalization of the last layer of the embeddings and of the outcome models.  

The training hyper-parameters include the initial learning rate, the batch size and the number of epochs. Training is conducted using  Adam \citep{kingmaAdamMethodStochastic2014}, with exponential weight decay. 
Overfitting is prevented using a validation set $\DV$ including  $30\%$ of the training set, and using the factual error to select the model after a fixed number of epochs. 

The propensity estimate is trained using mainstream supervised learning; logistic regression, k-nearest neighbors, and decision trees are considered. The model retained is determined by grid-search using a cross-validation scheme for each considered dataset.

\subsection{Experimental results \label{sec:experimentalvalidation}}
The average performance of \XX\ on all \IHDP\ instances is reported on \cref{tab:resultsIHDP}, compared with all other baselines\footnote{Results from other models taken from \cite{shalitEstimatingIndividualTreatment2017} or their respective articles. $\textit{OLS}_2$ stands for a \textit{T-learner} based on ordinary least square regression.}. On this dataset, \XX\ ranks first on within-sample \PEHE\ and second on out-of-sample \PEHE. Indeed, \XX\ is designed to optimize the \PEHE\ performance indicator, as motivated by \cref{theorem}. The comparatively lesser performance regarding within-sample and out-of-sample $\epsilon_{\ATE}$ is blamed on the $L_2$ regularization (more in \cref{sub:bias}), tending to bias the estimates toward 0, as noted by \cite{laanTargetedLearningCausal2011}.

\begin{figure}
\centering
\begin{subfigure}{.35\textwidth}
  \centering
  \includegraphics[width=1.\linewidth]{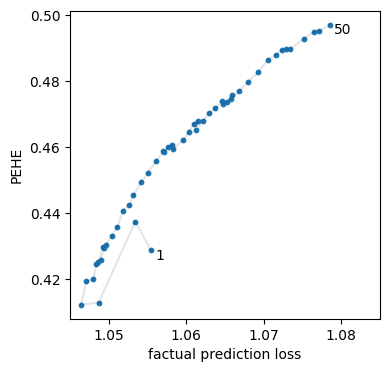}
  \caption{$\htau^{\textit{top-K}}$, $K \in\llbracket 1,50 \rrbracket$, \IHDP}
\end{subfigure}
\begin{subfigure}{.35\textwidth}
  \centering
  \includegraphics[width=1.\linewidth]{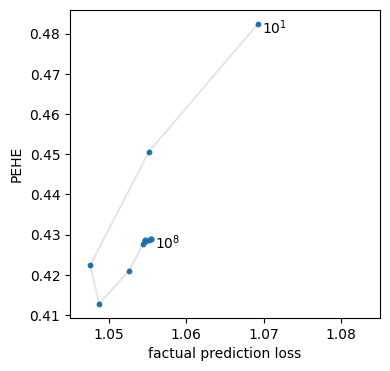}
  \caption{$\textit{softmax}_{\lambda}$, $\lambda\in\{10^{\nicefrac{k}{2}}\}_{k=2}^{16}$, \IHDP}
\end{subfigure}

\begin{subfigure}{.35\textwidth}
  \centering
  \includegraphics[width=1.\linewidth]{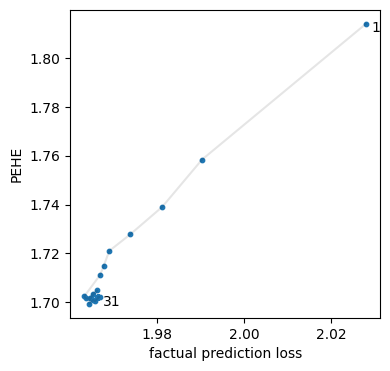}
  \caption{$\htau^{\textit{top-K}}$, $K\in\llbracket 1,31 \rrbracket$, \ACIC}
\end{subfigure}
\begin{subfigure}{.35\textwidth}
  \centering
  \includegraphics[width=1.\linewidth]{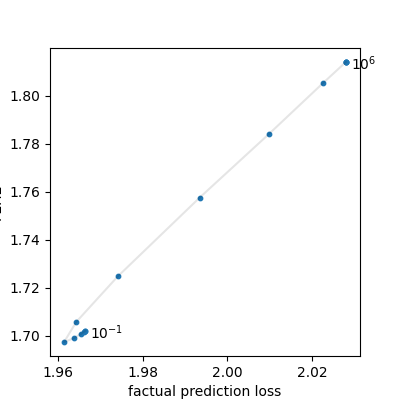}
  \caption{$\textit{softmax}_{\lambda}$, $\lambda\in\{10^{\nicefrac{k}{2}}\}_{k=\text{-}2}^{12}$, \ACIC}
\end{subfigure}

\centerline{~}
\caption{Ensemble models in the \textit{out-of-sample} setting: \PEHE\ vs factual error on the validation set sensitivity, depending on the hyper-parameter.} 
\label{fig:ensemblefiguresACIC}
\end{figure}

As could be expected \cite{breimanRandomForests2001}, ensemble variants of \XX\ introduced in \cref{sec:algo} improve on \XX. The \PEHE\ and (factual) accuracy depending on their hyperparameters (number $K$ of models in the top-$K$ ensemble, temperature $\lambda$ for the softmax weighted combination) are reported on Fig. \ref{fig:ensemblefiguresACIC} (one point per value of the hyperparameter). 

This figure suggests that the factual accuracy 
can reliably be used to select the hyperparameter ($K=4$ for the top-K ensemble and $\lambda=100$ for the softmax ensemble) yielding the best \PEHE. 
For these selected hyperparameters, 
the \PEHE\ is statistically significantly better than \XX\ with \textit{p-value}$=1.3e-3$ on a one-sided paired \textit{t-test}.

\begin{center}
\begin{table}
    \centering
    \begin{tblr}{
      hline{1,2,3} = {2-5}{solid},
      hline{4-25} = {1-5}{solid},
      vline{1} = {4-24}{solid},
      vline{2,6} = {1-24}{solid},
      vline{4} = {2-24}{solid},
      vline{3,5} = {3-24}{solid},
      cells = {c},
      cell{1}{2} = {c=4}{c},
      cell{2}{2} = {c=2}{c},
      cell{2}{4} = {c=2}{c},
      cell{2-20}{1} = {l},
      vspan = even,
    }
      & \textbf{\IHDP}  \\
      & \textit{within-sample} & & \textit{out-of-sample} \\
      & $\sqrt{\PEHE}$ & $\epsilon_\ATE$ & $\sqrt{\PEHE}$ & $\epsilon_\ATE$ \\
      $\textit{OLS}_2$\hfill\cite{shalitEstimatingIndividualTreatment2017} & $2.4 \pm .1$ & $.14 \pm .01$ & $2.5 \pm .1$ & $.31 \pm .02$ \\
      \textit{BART}\hfill\cite{hillBayesianNonparametricModeling2011a} & $2.1 \pm .1$ & $.23 \pm .01$ & $2.1 \pm .1$ & $.34 \pm .02$ \\
      \textit{BNN}\hfill\cite{johanssonLearningRepresentationsCounterfactual2016a} & $2.2 \pm .1$ & $.37 \pm .03$ & $2.1 \pm .1$ & $.42 \pm .03$ \\ 
      \textit{CF}\hfill\cite{wagerEstimationInferenceHeterogeneous2018} & $3.8 \pm .2$ & $.18 \pm .01$ &  $3.8 \pm .2$ & $.40 \pm .03$ \\ 
      \textit{CFR-Wass}\hfill\cite{shalitEstimatingIndividualTreatment2017} & $.71 \pm .0$ & $.25 \pm .01$ & $.76 \pm .0$ & $.27 \pm .01$ \\ 
      \textit{CEVAE}\hfill\cite{louizosCausalEffectInference2017c} & $2.7 \pm .1$ & $.34 \pm .01$ & $2.6 \pm .1$ & $.46 \pm .02$  \\ 
      \textit{SITE}\hfill\cite{yaoRepresentationLearningTreatment2018c} & $.60 \pm .09$ & \slash & $.66 \pm .11$ & \slash  \\ 
      \textit{GANITE}\hfill\cite{yoonGANITEEstimationIndividualized2018} & $1.9 \pm .4$ & $.43 \pm .05$ & $2.4 \pm .4$ & $.49 \pm .05$ \\ 
      \textit{NSGP}\hfill\cite{alaaLimitsEstimatingHeterogeneous2018c} & $.51 \pm .01$ & \slash & $.64 \pm .03$ & \slash \\ 
      \textit{ACE}\hfill\cite{yaoACEAdaptivelySimilarityPreserved2019} & $.49 \pm .005$ & \slash & $.54 \pm .06$ & \slash \\ 
      \textit{DKLITE}\hfill\cite{zhangLearningOverlappingRepresentations2020} & $.52 \pm .02$ & \slash & $.65 \pm .03$ & \slash \\ 
      \textit{DR-CFR}\hfill\cite{hassanpourLearningDisentangledRepresentations2019} & \slash & \slash & $.65 \pm .03$ & $\textbf{.03} \pm .04$ \\ 
      \textit{BWCFR}\hfill\cite{assaadCounterfactualRepresentationLearning2021a} & \slash & \slash & $.63 \pm .01$ & $.19 \pm .01$ \\ 
      \textit{ABCEI}\hfill\cite{duAdversarialBalancingbasedRepresentation2021a} & $.71 \pm .0$ & $\textbf{.09} \pm .01$ & $.73 \pm .0$ & $.09 \pm .01$ \\ 
      \textit{CBRE}\hfill\cite{zhouCycleBalancedRepresentationLearning2021} & $.52 \pm .0$ & $.10 \pm .01$ & $.60 \pm .1$ & $.13 \pm .02$ \\ 
      \textit{MIM-DRCFR}\hfill\cite{chengLearningDisentangledRepresentations2022} & \slash & \slash & $\textbf{.38} \pm .009$ & $.09 \pm .001$ \\
      \textit{DeR-CFR}\hfill\cite{wuLearningDecomposedRepresentations2023} & $.44 \pm .02$ & $.13 \pm .02$ & $.53 \pm .07$ & $.15 \pm .02$ \\    
      \textit{DRCFR+}\hfill\cite{chauhanAdversarialDeconfoundingIndividualised2023a} & $.86 \pm .01$ & \slash & $1.19 \pm .01$ & \slash \\
      \textbf{\XX} & $.42 \pm .03$ & $.12 \pm .009$ & $.43 \pm .02$ & $.13 \pm .01$ \\ 
      \textbf{\textit{K-top} \XX} & $\textbf{.40} \pm .02$ & $.10 \pm .008$ & $.41 \pm .02$ & $.11 \pm .009$ \\ 
      \textbf{$\textit{softmax}_{\lambda}$ \XX} & $\textbf{.40} \pm .023$ & $.095 \pm .008$ & $.42 \pm .02$ & $.11 \pm .008$ \\ 
    \end{tblr}
    \caption{\IHDP: Comparative performances of all baselines and \XX\ (lower is better). 
Statistically significantly better results are indicated in bold.    
    \label{tab:resultsIHDP}}
\end{table}
\end{center}

Likewise, the average performance of \XX\ on all \ACIC\ instances is reported on \cref{tab:resultsacic}, compared with all other baselines. On this dataset, \XX\ ranks second
for \PEHE\ both within-sample and out-of-sample, behind TEDVAE \cite{zhangTreatmentEffectEstimation2021}. A tentative interpretation for this fact 
is that
some problem instances use a uniform treatment assignment, while other problem instances are truly unbalanced. However, by design \XX\ is not suited to uniform treatment assignment: in such cases, it is not necessary to consider the counter-factualization of control and treatment samples independently. On the contrary, learning two separate pipelines, \Po\ and \Pz\, is likely to lead to over-fitting. The issue is all the more severe as the selected hyper-parameter setting is the same for all \ACIC\ instances. 

Most interestingly however, both 
ensemble variants of \XX\ significantly improve on 
\XX\ and on TEDVAE. It is reminded that the hyperparameters of the ensemble are selected (one value for all problem instances) as detailed 
in Appendix \ref{chapter:hyperparameter}.

\begin{center}
\begin{table}
    \centering
    \begin{tblr}{
      hline{1,2} = {2-3}{solid},
      hline{3-21} = {1-3}{solid},
      vline{1} = {3-20}{solid},
      vline{2,4} = {1-20}{solid},
      vline{3} = {2-20}{solid},
      cells = {c},
      cell{1}{2} = {c=2}{c},
      vspan = even,
    }
      & \textbf{\ACIC}  \\
      & \textit{within-sample} & \textit{out-of-sample} \\
        \textit{CF}\hfill\cite{atheyRecursivePartitioningHeterogeneous2016}  & 2.16 $\pm .17$ & 2.18 $\pm .18$ \\
        \textit{BART}\hfill\cite{hillBayesianNonparametricModeling2011a} & 2.13 $\pm .18$ & 2.17 $\pm .15$ \\
        \textit{X-learner(RF) }\hfill\cite{kunzelMetalearnersEstimatingHeterogeneous2019} & 1.86 $\pm .15$ & 1.89 $\pm .16$ \\
        \textit{CFR}\hfill\cite{shalitEstimatingIndividualTreatment2017} & 2.05 $\pm .18$ & 2.18 $\pm .20$ \\
        \textit{SITE}\hfill\cite{yaoRepresentationLearningTreatment2018c} & 2.32 $\pm .19$ & 2.41 $\pm .23$ \\
        \textit{DR-CFR}\hfill\cite{hassanpourLearningDisentangledRepresentations2019} & 2.44 $\pm .20$ & 2.56 $\pm .21$ \\
        \textit{GANITE}\hfill\cite{yoonGANITEEstimationIndividualized2018} & 3.12 $\pm .28$ & 3.28 $\pm .35$ \\
        \textit{CEVAE}\hfill\cite{louizosCausalEffectInference2017c} & 2.78 $\pm .56$ & 2.84 $\pm .61$ \\
        \textit{TEDVAE}\hfill\cite{zhangTreatmentEffectEstimation2021} & 1.75 $\pm .14$ & 1.77 $\pm .17$ \\
        \textbf{\XX} (ours) & 1.79 $\pm .11$ & 1.81 $\pm .13$ \\
        \textbf{\textit{K-top} \XX} & 1.69 $\pm .11$ & 1.70 $\pm .13$ \\ 
        \textbf{$\textit{softmax}_{\lambda}$ \XX} & 1.69 $\pm .11$ & 1.70 $\pm .13$ \\  
    \end{tblr}
    \caption{Comparative performances of \XX\ and baselines on \ACIC. All baseline performances are reported from \cite{zhangTreatmentEffectEstimation2021}. \label{tab:resultsacic}}
\end{table}
\end{center}

\subsection{Discussion \label{sub:bias}}
As we said, the comparatively poorer performance of \XX\ over \ATE, compared to \PEHE, is blamed on the regularization term (term $\gamma \|\Po\|$ of the training loss, Eq. \ref{eq:loss}), which aims to avoid overfitting. This interpretation is confirmed by complementary experiments \cite{lacom2024}.

\section{Conclusion}
The main contribution of the proposed approach lies in the observation that, when treatment assignment is not uniform, counterfactual estimation for control and treated samples induces two distinct problems. \XX\ takes this observation into account through an original neural architecture, hybridizing \xlearners\ and \tlearners.

Two quantities are thus defined to characterize the quality of a change of representation: the counter-factualizability of a sample, i.e. its distance from its twin in latent space; and the weight of the counterfactual importance of a sample $x$, counting how many samples admit $x$ as a twin. 
Finally, the models are learned by optimizing a learning loss that imposes a good change of representation: the aim is to ensure good counter-factualizability of all samples, and to ensure that the factual model is all the more accurate on examples with higher counterfactual importance weights. 

Learned models benefit from theoretical guarantees, noting that the upper bound of\PEHE\ involves the same terms as those involved in the learning loss. Last, the merits of the approach are experimentally demonstrated on the main two benchmarks of the domain, compared with the prominent approaches of the state of the art.
A word of caution: the approach is not suited to uniform treatment assignment: in this case, the two counterfactual estimation problems are in fact the same, and the two-pipeline \XX\ runs the risk of overfitting.

Several research perspectives are opened up by \XX. One short-term perspective is to jointly learn both pipelines, thus promoting their complementarity. 

Another perspective is to revise the notion of latent twin, which currently involves only the nearest neighbor of the sample under consideration. A more robust approach would be to consider several nearest neighbors of a sample in the latent space (e.g. using a Gaussian kernel) and revise the counterfactual importance weight of neighboring samples accordingly. 

A third perspective concerns the notion of uncertainty. The minimization of CATE uncertainty is at the heart of the proposed change of representation, which aims to strengthen the counter-factualizability of samples. An alternative could be based on the different models involved in the \XX\ ensemble, using them to assess uncertainty on factual and counterfactual estimates for any particular sample. 

Lastly, the extension of the approach to the multi-level treatment setting can be tackled, considering one pipeline per treatment level, plus one for the control setting.

\appendix

\section{Appendix}

\subsection{Proofs \label{appendix:proofs}}

\subsubsection[Proof of Theorem 1]{Proof of \cref{theorem}}

\begin{proof}
Let $(x,t,y)$ be a sample in $\cal D$ with $z = \phi(x)$, and let $(x', t'=1-t, y')$ be its mirror twin w.r.t $\phi$ ($\phi(x')=z'$). Assuming with no loss of generality that $t=1$, it comes:
\begin{align*}
    &|(\hnu^1\mm\hnu^0)(z) - (\nu^1\mm\nu^0)(z)|^2 \\
    &= |[\hnu^1(z)\mm\nu^1(z)] - [\hnu^0(z)\mm\hnu^0(z')] - [\hnu^0(z')\mm\nu^0(z')] - [\nu^0(z')\mm\nu^0(z)]|^2 \\
    &\leq 4 |\hnu^1(z)\mm\nu^1(z)|^2 + 4 |\hnu^0(z)\mm\hnu^0(z')|^2 + 4 |\hnu^0(z')\mm\nu^0(z')|^2 + 4 |\nu^0(z')\mm\nu^0(z)|^2 \\
    &\leq 4 |\hnu^1(z)\mm\nu^1(z)|^2 + 4 |\hnu^0(z')\mm\nu^0(z')|^2 + 4 (\widehat{L}^2+L^2) ||z'\mm z||^2 
\end{align*}
Averaging over $(x,t,y)$ in $\cal D$ yields the result.\label{proof} 
\end{proof}

\subsubsection[Proof of Theorem 2]{Proof of \cref{theorem2}}

\begin{proof}
Let $(x_i,t_i,y_i)$ be a sample in $\cal D$. Without loss of generality, assume $t_i=1$. Let $(x_j, t_j=0, y_j)$ be its mirror twin. Denote by $z_i$ and $z_j$ their respective representations: $\phio(x_i)=z_i,\phio(x_j)=z_j$. Using Cauchy-Schwarz applied on $\RR^5$, for any vector $u\in\RR^5$, 
$$(\sum u_i)^2\ = \langle \mathds{1},u\rangle^2  \leq ||\mathds{1}||^2\times||u||^2 = 5\sum u_i^2$$
with $\mathds{1}$ the vector with all coordinates set to 1. It then comes:
\begin{align*}
    \big(\Bar{\tau}_i-\tau(x_i)\big)^2
    =& \big((y_i- \HTCF(z_i)) - (\nuo^1-\nuo^0)(z_i)\big)^2 \\
    =& \big([y_i-\nuo^1(z_i)] + [\nuo^0(z_i)- \nuo^0(z_j)] + [\nuo^0(z_j)- y_j] \\
    &+ [y_j- \HTCF(z_j)]+ [ \HTCF(z_j)- \HTCF(z_i)]\big)^2 \\
    \leq& 5(y_i-\nuo^1(z_i))^2 + 5(\nuo^0(z_i)- \nuo^0(z_j))^2 + 5(\nuo^0(z_j)- y_j)^2 \\
    &+ 5(y_j- \HTCF(z_j))^2+ 5(\HTCF(z_j)- \HTCF(z_i))^2 \\
    \leq& 5(y_i-\mu^1(x_i))^2 + 5(y_j-\mu^0(y_j))^2 +  5(y_j- \HTCF(z_j))^2
    + 5 (\widehat{L}^2+L^2) ||z_j- z_i||^2
\end{align*}
Averaging over $(x_i,t_i=1)$ in $\cal D$, it comes:
\begin{align*}
    \frac{1}{n}\sum_{t_i=1}\big(\Bar{\tau}_i-\tau(x_i)\big)^2 
    \leq \frac{5}{n}& \sum_{t_i=1}\biggl(
    (L^2+\hL^2) \|\phi(x_i)-\phi(x_j)\|^2
    + \big(\mu^{1}(x_i)-y_i\big)^2 \biggl)\\
    + \frac{5}{n}& \sum_{t_j=0}\bigg(w_j \big[\big(\hoz\circ\phi_{1}(x_j)-y_j\big)^2
    + \big(\mu^{0}(x_j)-y_j\big)^2\big] \bigg)
\end{align*}
Adding the control samples sum yields $M_2$.
\end{proof}

\subsubsection[Proof of Theorem 3]{Proof of \cref{theorem3}}

\begin{proof}
Set the loss hyper-parameters values $\alphaz,\alphao,\betaz,\betao$ to 
\begin{align*}
    \big(\alphaz, \alphao, \betaz, \betao\big) &= \big((1-p)(L^2+\hL^2), p(L^2+\hL^2), 1, 1\big)\\
    \textit{entailing} \\
    \big( \frac{\alphaz}{n_0}, \frac{\alphao}{n_1}, \frac{\betaz}{n_1+\betaz n_0}, \frac{\betao}{n_0+\betao n_1} \big) &= \big( \frac{L^2+\hL^2}{n}, \frac{L^2+\hL^2}{n}, \frac{1}{n}, \frac{1}{n} \big)
\end{align*}

Then, the total loss over both pipelines total loss writes
{\allowdisplaybreaks
\begin{align*}
    \Lz\text{+}\Lo =& \Bigg(\frac{L^2+\hL^2}{n} \sum_{t_i=0} \|\phi_0(x_i)-\phi_0(x_i^{m,\phi_{0}})\|^2 + \frac{1}{n}\sum_{t_i=1}w_i(\HCCF\circ\phiz(x_i)\mm y_i)^2\Bigg)\\ 
    &+ \Bigg(\frac{1}{n(1\mm p)}\sum_{t_i=0}(\HCF\circ\phiz(x_i)\mm y_i)^2 + \frac{1}{n}\sum_{t_i=1}(\HCCF\circ\phiz(x_i)\mm y_i)^2 + \gammaz \Omega(\Pz)\Bigg)\\
    &+\Bigg(\frac{L^2+\hL^2}{n} \sum_{t_i=1} \|\phi_1(x_i)-\phi_1(x_i^{m,\phi_{1}})\|^2 + \frac{1}{n}\sum_{t_i=0}w_i(\HTCF\circ\phio(x_i)\mm y_i)^2\Bigg)\\ 
    &+ \Bigg(\frac{1}{np}\sum_{t_i=1}(\HTF\circ\phio(x_i)\mm y_i)^2 + \frac{1}{n}\sum_{t_i=0}(\HTCF\circ\phio(x_i)\mm y_i)^2 + \gammao \Omega(\Po)\Bigg)\\
    =& \frac{1}{n} 
    \sum_{t_i=0} \biggl(w_i \big(\hoz\circ\phi_{1}(x_i)\mm y_i\big)^2
    + (L^2+\hL^2) \|\phi_0(x_i)-\phi_0(x_i^{m,\phi_{0}})\|^2 \biggl)\\
    +& \frac{1}{n} 
    \sum_{t_i=1} \biggl(w_i \big(\hzo\circ\phi_{0}(x_i)\mm y_i\big)^2
    + (L^2+\hL^2) \|\phi_1(x_i)-\phi_1(x_i^{m,\phi_{1}})\|^2 \biggl)\\
    &+ \frac{1}{n}\sum_{i=1}^n(h_{1-t_i}^{t_i}\circ\phi_{1-t_i}(x_i)\mm y_i)^2 + \frac{1}{n}\sum_{i=1}^n \frac{(h_{t_i}^{t_i}\circ\phi_{t_i}(x_i)\mm y_i)^2}{t_i p + (1\mm t_i)(1\mm p)} + \gammaz\Omega(\Pz) + \gammao\Omega(\Po)\\
    =& \frac{1}{5}M_2 - \kappa_Y + \gammaz\Omega(\Pz) + \gammao\Omega(\Po) \\
    &+ \frac{1}{n}\sum_{i=1}^n(h_{1-t_i}^{t_i}\circ\phi_{1-t_i}(x_i)\mm y_i)^2
    + \frac{1}{n}\sum_{i=1}^n \frac{(h_{t_i}^{t_i}\circ\phi_{t_i}(x_i)\mm y_i)^2}{t_i p + (1\mm t_i)(1\mm p)} 
\end{align*}}
implying an upper bound on the within-sample empirical \PEHE:
\begin{align*}
    \frac{1}{n}\sum_{i=1}^n \big(\Bar{\tau}_i-\tau(x_i)\big)^2 &\leq 5\big(\Lz+\Lo\big) - 5\big(\gammaz\|\Pz\|^2 + \gammao\|\Po\|^2\big) + 5\kappa_Y \\
    &- \frac{5}{n}\sum_{i=1}^n \frac{1}{p^{t_i}}(h_{t_i}^{t_i}\circ\phi_{t_i}(x_i)-y_i)^2 - \frac{5}{n}\sum_{i=1}^n(h_{1-t_i}^{t_i}\circ\phi_{1-t_i}(x_i)-y_i)^2
\end{align*}    
\end{proof}

\subsubsection[Proof of lemma 4]{Proof of \cref{lemma:minimizing}}

\begin{proof}
    Let $\nu^{0*}$ be a minimizer of ${\cal L}(\cdot) = \EE_{X,Y|T=0}\big[\big(\cdot\circ\phi(X)-Y^0\big)^2|T=0\big]$. Then, $\nu^{0*}:z\in\mathcal{Z}\mapsto \EE[Y^0|T=0,\phi(X)=z]$ and therefore $\nu^{0*}\circ\phi: x\in\mathcal{X}\mapsto \EE[Y^0|T=0,\phi(X)=\phi(x)]$. The problem rephrases in: does the equality of $\EE[Y^0|T=0,\phi(X)=\phi(\cdot)]$ and $\mu^0(\cdot)$ hold? \\

    Let us first show that the average value of the control outcome $Y^0$ conditionally on $X$ equals the average value conditionally on $\phi(X)$. Let then $x$ be an element of $\mathcal{X}$. Denote by $U$ the set of events $\phi^{-1}(\{\phi(x)\})$. The formula of total probability writes
    \begin{align*}
        \EE[Y^0|\phi(X)=\phi(x)] 
        &= \int\nolimits_{U} \EE[Y^0|\phi(X)=\phi(x),X=u] \PP(X=u|\phi(X)=\phi(x))\textit{du} \\
        &= \int\nolimits_{U} \EE[Y^0|X=u] \PP(X=u|\phi(X)=\phi(x))\textit{du} \\
        &= \int\nolimits_{U} \mu^0(u) \PP(X=u|\phi(X)=\phi(x))\textit{du} \\
        &= \int\nolimits_{U} \nu^0\circ\phi(u) \PP(X=u|\phi(X)=\phi(x))\textit{du} \\
        &= \int\nolimits_{U} \nu^0\circ\phi(x) \PP(X=u|\phi(X)=\phi(x))\textit{du} \\
        &= \nu^0\circ\phi(x) \int\nolimits_{U} \PP(X=u|\phi(X)=\phi(x))\textit{du} \\
        &= \mu^0(x)
    \end{align*}

    Now let us show that $\EE[Y^0|\phi(X)=\phi(\cdot)]$ and $\EE[Y^0|T=0,\phi(X)=\phi(\cdot)]$ are also equal. Denote by $V$ the set of events $(\phi(X),T)^{-1}(\{(\phi(x),0)\})$. Then,
    \begin{align*}
        \EE[Y^0|\phi(X)=\phi(x),T=0]
        &= \int\nolimits_{ V} \EE[Y^0|\phi(X)=\phi(x),T=0,X=v]\PP(X=v|\phi(X)=\phi(x),T=0)\textit{dv} \\
        &= \int\nolimits_{ V} \EE[Y^0|T=0,X=v]\PP(X=v|\phi(X)=\phi(x),T=0)\textit{dv} \\
        &= \int\nolimits_{ V} \EE[Y^0|X=v]\PP(X=v|\phi(X)=\phi(x),T=0)\textit{dv}\\
        &= \int\nolimits_{ V} \EE[Y^0|\phi(X)=\phi(x)]\PP(X=v|\phi(X)=\phi(x),T=0)\textit{dv}\ \hspace{20pt} \textit{(cf. above)}\\
        &= \EE[Y^0|\phi(X)=\phi(x)]\int_{ V}\PP(X=v|\phi(X)=\phi(x),T=0)\textit{dv} \\
        &= \EE[Y^0|\phi(X)=\phi(x)]
    \end{align*}

    Finally, $\EE[Y^0|T=0,\phi(X)=\phi(\cdot)] = \mu^0(\cdot)$, and $\nu^0$ minimizes $\mathcal{L}$. 
\label{proof:lemma4}
\end{proof}

\subsection[Additional remarks regarding the discussion of the assumptions]{Additional remarks regarding the discussion of the formal analysis \pcref{sec:alrite-discu}}

Although the main result of \cite{shalitEstimatingIndividualTreatment2017} resorts to the invertibility of the mapping $\phi$, strictly weaker assumptions allow for $\nu^0$ to reach the minimum of $\mathcal{L}$. In particular, if $\phi$ is injective, then conditioning on $X=x$ is equivalent to conditioning on $\phi(X)=\phi(x)$, and the equality $\EE[Y^0|T=0,\phi(X)=\phi(\cdot)] = \mu^0(\cdot)$ directly follows. \\

The equality still holds\footnote{
Assume that $\phi(X)$ is sufficient with respect to $Y^0$: $Y^0\indep X|\phi(X)$ (or equivalently $\phi(X)$ is a prognostic score). Then, 
\begin{align*}
    \mu^0(x) &= \EE[Y^0|X=x] \\
    &= \EE[Y^0|\phi(X)=\phi(x),X=x] \\
    &= \EE[Y^0|\phi(X)=\phi(x)]
\end{align*}
As such, $\mu^0(x)$ is entirely determined by $\phi(x)$; the existence of $\nu^0$ is guaranteed, and the result derives from \cref{lemma:minimizing}.
\label{footnote}}
in the \textbf{strictly weaker} setting where $\phi(X)$ is a sufficient statistic with respect to $Y^0$, in the sense that $Y^0 \indep X | \phi(X)$. \\

Finally, as shown by \cref{lemma:minimizing}, the equality still holds under the \textbf{strictly weaker}\footnote{
$\phi(X)$ being a sufficient statistic with respect to $Y^0$ implies the existence of such $\nuo$ according to \cref{footnote} \\

To illustrate that this is no necessary condition, consider the following counter-example. Suppose that $\mathcal{X} = [0,1] \times [0,1]$, with samples being drawn uniformly in $\mathcal{X}$, and $Y^0\sim\mathcal{N}(X_1, X_2^2)$. Let $\phi:(x_1,x_2) \mapsto x_1$ be the projection on the first feature axis. \\

Since $\EE[Y^0|X=x] = \phi(x)$, $\nu^0 = \textit{Id}$ verifies the condition. However, $\phi(X)$ being fixed, the variance of $Y^0$ depends on $X_2$; $Y^0\notindep X|\phi(X)$ and $\phi(X)$ is no sufficient statistic for $Y^0$.} 
hypothesis that there exists $\nu^0$ such that $\mu^0=\nu^0\circ\phi$. \\

Note that the hypothesis $Y^0 \indep T | \phi(X)$ isn't sufficient\footnote{
Even if $\EE[Y^0|T=0,\phi(X)=\phi(x)] = \EE[Y^0|\phi(X)=\phi(x)]$, there is no reason for $\EE[Y^0|\phi(X)=\phi(x)]$ to be equal to $\mu^0(x)$.
} 
nor necessary\footnote{
Consider the following setting:
\begin{equation}
        \begin{cases}
        \begin{aligned}
            X &\sim (\textit{Ber}(\nicefrac{1}{2}),\textit{Ber}(\nicefrac{1}{2})) \\
            \phi(X) &= X_1 \\
            T &= X_2\times(2n_T-1) + 1-n_T, &n_T\sim\textit{Ber}(.9) \\
            \YC &= (X_1-\nicefrac{1}{2}) + (X_2+\nicefrac{1}{2})\times n_Y, &n_Y\sim\mathcal{N}(0,.1)
        \end{aligned}
        \end{cases}
    \label{remarkinsidenr}
\end{equation}
Set now $\nu^0:z\mapsto z-\nicefrac{1}{2}$. Then, $\EE[Y^0|X=x]=x_1-\nicefrac{1}{2}=\nu^0\circ\phi(x)$. Conditional exchangeability w.r.t. $X$, positivity and \SUTVA\ hold. However, $Y^0 \notindep T | \phi(X)$. $X_1$ being fixed, the variance of $Y^0$ is larger when $T$ takes value 1 than when it takes value 0: $\YC\notindep T|\phi(X)$.
}.

\subsubsection{Asymptotical behavior of \xxx \label{appendix:asymptotical}}
Given a mapping $\phi$ and in the large sample limit, the \xxx\ of any sample goes to 0 in probability.

\begin{proof}
Let us assume for simplicity that $\mathcal{X} \subset \RR^d$. Let $(x_i,t_i=0,y_i)$ be a control sample from the training dataset $\mathcal{D}$, with $\epsilon > 0$. 
Function $\phiz$, being implemented with a finite-weights neural network, is continuous. As such, the inverse image of the latent space open ball $\textit{B}(\phiz(x_i),\epsilon)$ centered on $\phiz(x_i)$ with radius $\epsilon$ is also an open. 
$(x_i,t_i=0,y_i)$ has been sampled from $\PP_{X,T,Y}$ and belongs to $\phiz^{-1}(\textit{B}(\phiz(x_i),\epsilon))$ so there also exists an open $A \subset \phiz^{-1}(\textit{B}(\phiz(x_i),\epsilon))$ of $\RR^d$ such that $\PP(T=0, X\in B)>0$. Since positivity holds, $\PP(T=1, X\in A)>0$. Finally, 
\begin{align*}
    \PP\big(\|\phi_0(x_i)-\phi_0(\mti)\| >\epsilon \big)
    &= \PP\big( \forall j\in\onen,\ t_j=1\ \implies\ \phiz(x_j)\not\in\ \textit{B}(\phiz(x_i),\epsilon) \big) \\
    &\leq \PP\big( \forall i\in\onen,\ t_j=1\ \implies\ x_j\not\in\ \phiz^{-1}(\textit{B}(\phiz(x_i),\epsilon)) \big) \\
    &\leq \PP\big( \forall j\in\onen,\ t_j=1\ \implies\ x_j\not\in\ A \big) \\
    &\leq \big(1- \PP(T=1, X\in A) \big)^n \\
    &\xrightarrow{n\rightarrow +\infty} 0
\end{align*}    
\end{proof}

\section{Hyper-parameter selection in causal inference \label{chapter:hyperparameter}}
\XX\ hyper-parameters are summarized in Table \ref{tab:chosenhyperparams}, together with their range of variation and the selected values for benchmarks \IHDP\ and \ACIC. 

\begin{table}[ht]
    \centering
    \begin{tblr}{
    hline{1} = {3-6}{solid},
    hline{2} = {2-6}{solid},
    hline{3-10} = {1-6}{solid},
    vline{1} = {3-9}{solid},
    vline{3,5,7} = {1-9}{solid},
    vline{2,4,6} = {2-9}{solid},
    cell{1}{3,5} = {c=2}{c},
    cells = {c},
    cell{-}{1} = {font=\small},
    cell{1}{-} = {font=\bfseries},
    hspan = even,
    }
    &  & \IHDP &  & \ACIC & \\
    & Range & \Pz & \Po &   \Pz & \Po \\
    regularization strength $\alpha$ & $\{0\} \cup\ \{10^{k/2}\}_{k=-4}^{4} $ & $10^0$ & $10^{1.5}$ & $10^{-2}$ & $10^{-2}$  \\
    reweighting importance $\beta$ & $\{0\} \cup\ \{10^{k/2}\}_{k=-4}^{2}$ & $10^{-1.5}$ & $0$ & $10^0$ & $0$ \\
    embedding model layers & $\llbracket 1, 5 \rrbracket$ & $4$ & $4$ & $5$ & $5$ \\
    outcome model layers    & $\llbracket 1, 5 \rrbracket$ & $3$ & $4$ & $5$ & $5$ \\
    embedding model width   & $\{ 20, 50, 100, 200\}$ & $20$ & $20$ & $100$ & $100$ \\
    outcome model width     & $\{ 20, 50, 100, 200\}$ & $50$ & $100$ & $100$ & $100$ \\
    batch size              & $\{ 50, 100, 200, 500\}$ & $200$ & $200$ & $250$ & $250$ \\
    \end{tblr}
    \caption{\XX: Range of variation for the hyper-parameters, and selected hyper-parameters values. \label{tab:chosenhyperparams}}
\end{table}

The selection of the  hyper-parameter setting,  referred to as the AutoML problem in the mainstream supervised learning framework \citep{hutterAutomatedMachineLearning2019a}, 
is all the more severe in the CATE framework as the counterfactual information is \textit{de facto} unknown; it thus prevents the usual performance indicators (\PEHE) from being computed on any validation set $\DV$ sampled from the observational dataset $\cal D$.

The state of the art mostly relies on a  model-dependent methodology, e.g., 
 \cite{atheyRecursivePartitioningHeterogeneous2016} for causal trees, \cite{powersMethodsHeterogeneousTreatment2018} for causal boosting and bagged causal multivariate adaptive regression splines, or \cite{alaaBayesianNonparametricCausal2018} for Gaussian processes.

The approach proposed in this paper aims to be model-agnostic. It builds upon:  i) defining a proxy metric that can be computed on the validation set; ii)  retaining the model optimizing the considered proxy metric. The proxy metric thus plays the same role as a surrogate \PEHE. Following \citep{shalitEstimatingIndividualTreatment2017}, many authors rely on using $\taurisk_{\textit{1NNI}}$ as a proxy metric (detailed below). However, the 1-nearest neighbor estimator is known for its poor performance in middle to high-dimensional settings.\footnote{Furthermore, there exists evidence for 1NNI failure on synthetic problems \cite{schulerComparisonMethodsModel2018}.}

A set of proxy metrics has thus been considered (Table \ref{tab:feasible}), and these metrics have been compared along several performance indicators to select the most robust one.

\subsection{Proxy metrics}
We distinguish: 
\begin{itemize}
    \item the $\mu$\textbf{-risks}, only measuring the (possibly reweighted) factual validation prediction error;
    \item the \textbf{R-risk}, based on the `Robinson's transformation' \citep{robinsonRootNConsistentSemiparametricRegression1988} \pcref{eq:robinson}, that approximates $\EE\big[\big(Y -m(X) - (T-\eta(X))\htau(X)\big)^2\big]$ through approximating the mean conditional outcome $m(x) = \EE[Y|X=x]$ and the propensity $\eta$;
    \item the $\tau$\textbf{-risks}, leveraging an independent approximation of the ground-truth \CATE, that is most often based on a plug-in referred to as {One Nearest-Neighbor Imputation} (1NNI) \citep{shalitEstimatingIndividualTreatment2017, duAdversarialBalancingbasedRepresentation2021a, zhouCycleBalancedRepresentationLearning2021}, where the counter-factual outcome of $x$ is approximated by the factual outcome of its closest neighbor with opposite treatment assignment in instance space, noted $\overline{(x,t,y)}$. 
\end{itemize}

\begin{table}[ht]
    \begin{tblr}{
      hlines,
      vlines,
      column{1}={.13\textwidth},
      column{2}={.2\textwidth},
      column{3}={.55\textwidth},
      hspan=minimal,
    }
    \textbf{Estimator} & \textbf{Motivation} & \textbf{Expression:} $\frac{1}{|\DV|}\sum_{(x,t,y)\in\DV}\bullet$ \\ \hline
    $\widehat{\murisk}$ & factual error & $\big(y-\hmu^{t}(x)\big)^2$ \\
    $\widehat{\murisk}_\textit{IPTW}$ & same + \textit{IPTW} & $\hrho^{t}(x)\big(y - \hmu^{t}(x)\big)^2$  \\
    $\widehat{\textit{R-risk}}$ & \rlearners & $\Big(\htau(x)\big(t-\heta(x)\big) - \big(y-\hat{m}(x)\big)\Big)^2$ \\
    $\widehat{\taurisk}_{\textit{naive}}$ & simplicity & $\big(\htau(x) - (\hmu^1(x)-\hmu^0(x))\big)^2$ \\
    $\widehat{\taurisk}_{\textit{1NNI}}$ & \textit{1NNI} & $\big(\htau(x) - (2t-1)(y-\overline{(x,t,y)})\big)^2, $ \\
    $\widehat{\taurisk}_{\textit{IPTW}}$ & \flearners & $\big(\htau(x) - (2t-1) \hrho^{t}(x) y\big)^2$ \\ 
    $\widehat{\taurisk}_{\textit{U}}$ & \ulearners & $\big(\htau(x) -(2t-1)\hrho^{1-t}(x)\big(y-\hat{m}(x)\big)\big)^2$ \\
    $\widehat{\taurisk}_{\textit{DR}}$ & double robustness & $\Big(\htau(x) -\big( (\hmu^1\mm\hmu^0)(x) + (2t-1) \hrho^{t}(x) \big(y-\hmu^{t}(x)\big) \big)\Big)^2$  
    \end{tblr}
    \caption{Proxy metrics; $\hrho$ denotes the inverse of the approximate propensity $\heta$. \label{tab:feasible}}
\end{table}

Some of the above-mentioned proxy metrics may rely on auxiliary models that provide independent estimates of the outcome functions $\mu^0$, $\mu^1$, mean outcome $m$ and the propensity $\eta$. \\
Models $\hmu^0,\hmu^1,\hat{m}$ are enforced by Nu-Support Vector Regressors (\textit{NuSVR}) models \citep{plattProbabilisticOutputsSupport2000}, since they achieve low high cross-validation factual prediction loss. \\
Propensity estimators are logistic regressions, k-nearest neighbors, or decision tree regressors. These estimators and their hyper-parameters are chosen through a classical cross-validation procedure. 

\subsection{Selecting a proxy metrics \label{sub:proxy}}
On each benchmark, two sets of respectively $\ell_{\PZ}$ and $\ell_{\PO}$ hyper-parameter settings are randomly sampled, defining a total of $C = \ell_{\PZ} \times \ell_{\PO}$ candidate estimates 
 $$\htau^{(\ell,\ell')} = (1-\heta) \htauz^{(\ell)} + \heta\ \htauo^{(\ell')}$$

For each $j$-th candidate model ($j=1 \ldots C$), its \PEHE\ on the validation set noted $u_j$ is compared with the value of the proxy metrics, noted $v_j$. The reliability of the proxy metrics is assessed along three performance indicators: the Spearman correlation;\footnote{The {Spearman correlation} coefficient $\rho_s(u,v)$ is defined as the Pearson correlation coefficient between the rank vectors $r(u)=(r_u(1),\dots,r_u(C))$ and $r(v)=(r_v(1),\dots,r_v(C))$:
    \begin{equation}
        \rho_s(u,v) = \frac{\textit{Cov}\big(r(u),r(v)\big)}{\sigma\big(r(u)\big)\sigma\big(r(v)\big)}
    \end{equation}} the Kendall rank correlation\footnote{The Kendall rank correlation is proportional to the fraction of pairs ($1 \le i \le j \le C$ such that $(u_i, u_j)$ and $(v_i,v_j)$ are similarly ordered.} \citep{kendallNewMeasureRank1938}; and the discounted cumulative gain (DCG)\footnote{Assuming with no loss of generality that the models are ordered by increasing \PEHE, with $m(j)$ the rank of the $j$-th model according to the proxy metrics (only the first $p$ items are considered), 
    \begin{equation*}
        \DCG_p(u,v) = \sum_{j=1}^p \frac{2^{m(j)}}{\log(j+1)}
    \end{equation*}
    } \citep{jarvelinCumulatedGainbasedEvaluation2002}. 
    
\begin{figure}[htbp]

\begin{subfigure}[t]{0.2\textwidth}
    \includegraphics[width=\linewidth]{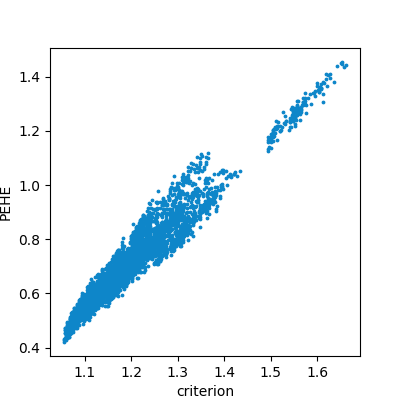}
\caption{\murisk}
\end{subfigure}\hfill
\begin{subfigure}[t]{0.2\textwidth}
  \includegraphics[width=\linewidth]{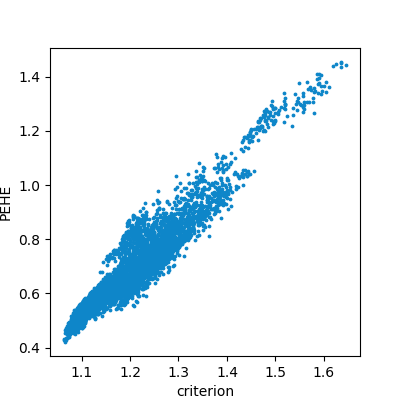}
\caption{$\murisk_{\textit{IPTW}}$}
\end{subfigure}\hfill
\begin{subfigure}[t]{0.2\textwidth}
    \includegraphics[width=\linewidth]{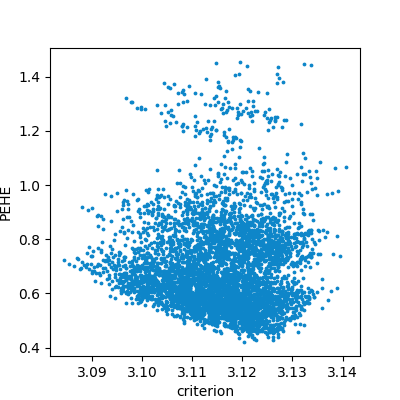}
\caption{\textit{R-risk}}
\end{subfigure}\hfill
\begin{subfigure}[t]{0.2\textwidth}
    \includegraphics[width=\linewidth]{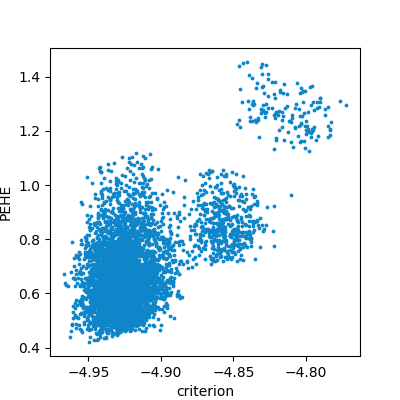}
\caption{$\pirisk_{\textit{IPTW}}$}
\end{subfigure}\hfill
\begin{subfigure}[t]{0.2\textwidth}
    \includegraphics[width=\linewidth]{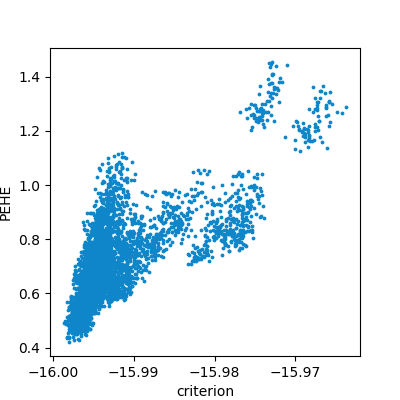}
\caption{$\pirisk_{\textit{DR}}$}
\end{subfigure}

\begin{subfigure}[t]{0.2\textwidth}
\includegraphics[width=\linewidth]{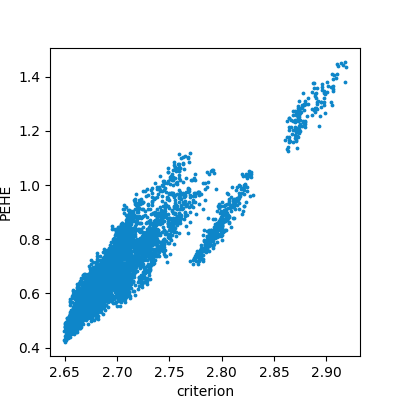}
\caption{$\taurisk_{\textit{1NNI}}$}
\end{subfigure}\hfill
\begin{subfigure}[t]{0.2\textwidth}
    \includegraphics[width=\linewidth]{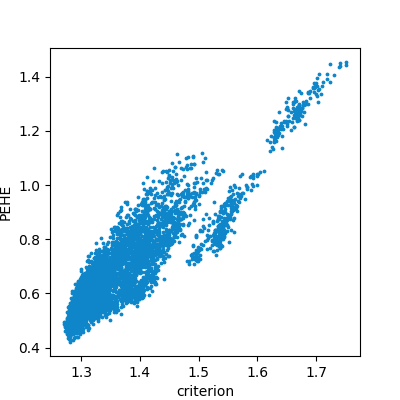}
\caption{$\taurisk_{\textit{naive}}$}
\end{subfigure}\hfill
\begin{subfigure}[t]{0.2\textwidth}
    \includegraphics[width=\linewidth]{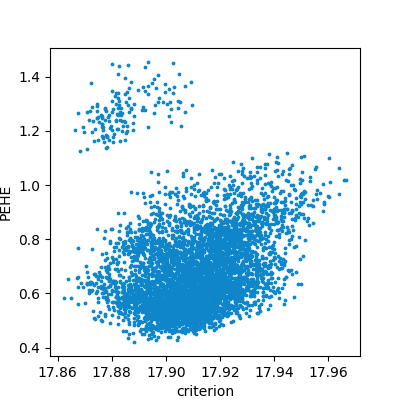}
\caption{$\taurisk_{\textit{IPTW}}$}
\end{subfigure}\hfill
\begin{subfigure}[t]{0.2\textwidth}
    \includegraphics[width=\linewidth]{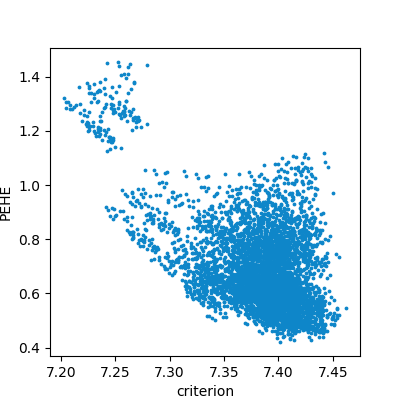}
\caption{$\taurisk_{\textit{U}}$}
\end{subfigure}\hfill
\begin{subfigure}[t]{0.2\textwidth}
    \includegraphics[width=\linewidth]{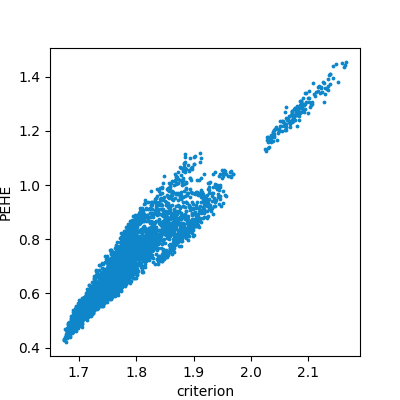}
\caption{$\taurisk_{\textit{DR}}$}
\end{subfigure}
\caption{\PEHE\ value (test set) vs proxy metrics (validation set) for \IHDP.}
\label{fig:manyscores}
\end{figure}

The values taken by the \PEHE\ and the proxy metrics  on the validation set of \IHDP\ are visually displayed in \cref{fig:manyscores}. 
Informally speaking, an ideal proxy metrics would show up as a diagonal line, displaying the identity of the true and the surrogate \PEHE, i.e. the proxy metrics. \cref{fig:manyscores} thus suggests that $\widehat{\murisk}$ and $\widehat{\taurisk}_{\textit{DR}}$ constitute the best proxy metrics.

Quantitatively,  the reliability of each proxy metrics is measured using Spearman correlation, Kendall rank correlation and DCG in \cref{tab:baselinemodelsreportedvalues}. 

We underline that this comparison can only be done in retrospect, for in real-life cases no evaluation of the \PEHE\ is possible.

\begin{table}
    \centering
    \begin{tblr}{
      hlines,
      vlines,
      cells = {l},
      cell{1}{-} = {font=\bfseries},
      hspan=even,
    }
    score & \textcolor{B4}{$\epsilon_{\ATE}$} & \textcolor{B4}{$\sqrt{\PEHE}$} & \textcolor{C2}{Spearman} & \textcolor{C2}{-\DCG} & \textcolor{C2}{Kendall} \\
    $\murisk$                   & 0.131 & \textbf{0.431} & \textbf{0.969} & -2.016 & \textbf{0.846} \\
    $\murisk_{\textit{IPTW}}$   & 0.131 & \textbf{0.431} & 0.942 & -2.014 & 0.798 \\
    $\pirisk_{\textit{IPTW}}$   & 0.139 & 0.672 & 0.366 & -2.751 & 0.253 \\
    $\pirisk_{\textit{DR}}$     & 0.136 & 0.490 & 0.796 & -2.239 & 0.598 \\
    $\textit{R-risk}$           & 0.146 & 0.723 & -0.064 & -3.304 & -0.049 \\
    $\taurisk_{\textit{naive}}$ & \textbf{0.125} & 0.496 & 0.858 & -2.196 & 0.668 \\
    $\taurisk_{\textit{1NNI}}$  & \textbf{0.125} & 0.462 & 0.903 & -2.070 & 0.729 \\
    $\taurisk_{\textit{IPTW}}$  & 0.141 & 0.581 & 0.282 & -3.281 & 0.199 \\
    $\taurisk_{\textit{U}}$     & 0.183 & 1.320 & -0.357 & -5.924 & -0.252 \\
    $\taurisk_{\textit{DR}}$    & 0.131 & \textbf{0.431} & 0.960 & \textbf{-1.985} & 0.825 \\
    \end{tblr}
    \caption{Hyper-parameter selection on \IHDP. Columns 2-3 display the \textcolor{B4}{performance of the model selected}  using the proxy metrics in column 1 (the lower the better). The associated  \textcolor{C2}{scores correlation metrics} are reported in columns 4-6  (the higher the better). \label{tab:baselinemodelsreportedvalues}}
\end{table}


These experiments validate \textit{a posteriori} the relevance of \murisk\ as best proxy metrics, as it achieves the lowest \PEHE\ and lowest \ATE. The quality of the proxy metrics is also confirmed by its high Spearman correlation coefficient, Kendall rank correlation coefficient, and discounted cumulative gain.


The proxy metrics empirically selected in \cite{shalitEstimatingIndividualTreatment2017}, $\taurisk_{\textit{1NNI}}$, also appears to be a reliable proxy, supporting a good model selection.

While $\murisk_{\textit{IPTW}}$, $\pirisk_{\textit{DR}}$, $\taurisk_{\textit{naive}}$ and $\taurisk_{\textit{DR}}$ seem worthy in retrospect, \textit{R-risk}, $\pirisk_{\textit{IPTW}}$, $\taurisk_{\textit{IPTW}}$ and $\taurisk_{\textit{U}}$ achieve poor selection performance.

\section{Ensemble \XX\ models \label{sec:ensemble}}
As mentioned \pcref{chapter:hyperparameter}, the hyper-parameter selection task relies on learning $\ell_0$ models $\htauz$ and $\ell_1$ models $\htauo$. Instead of selecting the best model according to the proxy metrics \pcref{sub:proxy}, we can return the best ensemble model combining the elementary models. Two types of combination have been considered in the experiments. 

\subsection{\textit{Top-K} ensemble}
With no loss of generality, let us assume that the $\htauz^{(i)}$ (resp. the $\htauo^{(j)}$) models are sorted by decreasing $\murisk$.
For $1 \le k \le \ell_0$, let $\overline{\htauz}^{k}$ be defined as the average of the first $k$ $\htauz$ models:
\[ \overline{\htauz}^{k}(x) = \frac{1}{k} \sum_{i=1}^k\htauz^{(i)}(x)\]
and let $\overline{\htauo}^{k}$ be likewise defined as the average of the first $k$ $\htauo$ models.
The overall ensemble $\overline{\htau}^{k}$, called top-K ensemble, is defined as:
\begin{equation}
    \overline{\htau}^{K}(x) = (1-\heta(x)) \,\overline{\htauz}^{K}(x)\, +\, \heta(x)\,\overline{\htauo}^{K}(x)
    \label{eq:topk}
\end{equation}
The hyper-parameter $K$, with $1 \le K \le \textit{min}(\ell_0,\ell_1)$, is likewise selected by optimizing the $\murisk$ of the ensemble. 

\subsection{$\lambda$ ensemble}
The second \XX\ ensemble, called $\lambda$ ensemble, is based on the weighted sum of the elementary models, where the weight of the $i$-th model is given as the softmax of the associated $\murisk$. 
Accordingly, the weighted sum of the $\htauz$ models, noted $\htauz^\lambda$, is defined as:
\[ \htauz^{\lambda}(x) = \frac{1}{\sum_{j=1}^{\ell_0} \exp\{-\lambda\times\murisk(\htauz^{j})\} } \hspace*{.1in} \sum_{i=1}^{\ell_0} \exp\{-\lambda\times\murisk(\htauz^{i}) \} \, \htauz^{i}(x)\]
with $\lambda > 0$. 
The weighted sum of the $\htauo$ models, denoted $\htauo^\lambda$, is likewise defined. Finally the overall ensemble $\htau^\lambda$ is defined as:

\begin{equation*}
    \htau^{\lambda}(x) =  (1-\heta(x))\, \htauz^\lambda(x) \,+ \,\heta(x) \,\htauo^\lambda(x)
\end{equation*}

Parameter $\lambda$ is selected by maximizing the $\murisk$ of the $\lambda$ ensemble.

\end{document}